\documentclass[]{jair}

\setcopyright{cc}
\copyrightyear{2026}
\acmDOI{10.1613/jair.1.19795}

\JAIRAE{Davide Grossi}
\JAIRTrack{} %
\acmVolume{86}
\acmArticle{11}
\acmMonth{06}
\acmYear{2026}

\RequirePackage[
  datamodel=acmdatamodel,
  style=acmauthoryear,
  backend=biber,
  giveninits=true,
  ]{biblatex}

\usepackage{booktabs}
\usepackage{subcaption}
\usepackage{caption}
\usepackage{amsmath}
\usepackage{mathtools}
\usepackage[shortcuts]{extdash}
\usepackage{siunitx}
\usepackage{enumitem}
\usepackage{bm}
\usepackage[capitalize,noabbrev]{cleveref}
\usepackage{mwe}
\usepackage{wrapfig}
\usepackage[all]{foreign}
\usepackage[acronym,nonumberlist]{glossaries}
\usepackage{adjustbox}

\newcommand{\sitp}{%
  \sbox0{$\lozenge$}%
  \usebox0\kern-.5\wd0\clap{{\scalebox{.7}[1]{$-$}}}\kern.5\wd0%
}

\DeclareSymbolFont{symbolsC}{U}{pxsyc}{m}{n}
\SetSymbolFont{symbolsC}{bold}{U}{pxsyc}{bx}{n}
\DeclareFontSubstitution{U}{pxsyc}{m}{n}
\DeclareMathSymbol{\medcirc}{\mathbin}{symbolsC}{7}

\hypersetup{
    pdfauthor={Olaf Lipinski, Adam J. Sobey, Federico Cerutti, Timothy J. Norman},
	pdftitle={It's About Time: Temporal References in Emergent Communication},
    pdflang={en-GB},
}

\begin{document}

\title[Temporal References in Emergent Communication]{It's About Time: Temporal References in Emergent Communication}

\received{09 July 2025}
\received[accepted]{19 March 2026}

\author{Olaf Lipinski}
\authornote{Corresponding Author.}
\email{o.lipinski@soton.ac.uk}
\orcid{0000-0002-2023-7617}
\affiliation{%
  \institution{University of Southampton}
  \city{Southampton}
  \country{United Kingdom}
}

\author{Adam J. Sobey}
\email{ajs502@soton.ac.uk}
\orcid{0000-0001-6880-8338}
\affiliation{%
  \institution{The Alan Turing Institute}
  \city{London}
  \country{United Kingdom}
}
\affiliation{%
  \institution{University of Southampton}
  \city{Southampton}
  \country{United Kingdom}
}

\author{Federico Cerutti}
\email{federico.cerutti@unibs.it}
\orcid{0000-0003-0755-0358}
\affiliation{%
  \institution{University of Brescia}
  \city{Brescia}
  \country{Italy}
}

\author{Timothy J. Norman}
\email{t.j.norman@soton.ac.uk}
\orcid{0000-0002-6387-4034}
\affiliation{%
  \institution{University of Southampton}
  \city{Southampton}
  \country{United Kingdom}
}

\renewcommand{\shortauthors}{Lipinski, Sobey, Cerutti \& Norman}

\glsdisablehyper
\newacronym{ec}{EC}{Emergent Communication}
\newacronym{pltl}{PLTL}{Past LTL}
\newacronym{ltl}{LTL}{Linear Temporal Logic}
\newacronym{trgs}{TRGs}{Temporal Referential Games}
\newacronym{trg}{TRG}{Temporal Referential Game}

\newcommand{\basen}{\textit{Base}\xspace}
\newcommand{\temporaln}{\textit{Temporal}\xspace}
\newcommand{\temporalrn}{\textit{TemporalR}\xspace}

\begin{abstract}
Emergent communication enables agents to develop bespoke languages that improve communication efficiency. Despite the known importance of temporal structure in natural language, there is no existing evidence of temporal references in emergent communication. This paper addresses this gap, by exploring how agents communicate about temporal relationships. We analyse three potential factors for the emergence of temporal references: environmental, external, and architectural. Our experiments demonstrate that altering the loss function is insufficient for temporal references to emerge; rather, architectural changes are necessary. A minimal change in agent architecture, using a different batching method, allows the emergence of temporal references. This modified design is compared with the standard architecture in a temporal referential games environment, which emphasises temporal relationships. The analysis shows that over 95\% of the agents with the modified batching method develop temporal references, without changes to their loss function. We consider temporal referencing necessary for future improvements to the agents' communication efficiency, enabling future agents to use a closer to optimal coding as compared to purely compositional languages. These insights provide the basis for incorporation of temporal references into other emergent communication settings, and investigation of other aspects of language.
\end{abstract}

\maketitle

\section{Introduction}\label{sec: introduction}

In emergent communication, autonomous agents develop their language from scratch. This contrasts with formally defined languages \citep{vieira_formal_2007}, where the agents are prescribed what and how to communicate \citep{rochlin_constraining_2015}. The emergent communication approach results in a vocabulary that is tailored to the specific environment in which the agents have been trained, reflecting the tasks the agents perform, the actions available to them and the other agents they interact with. These properties make the emergent language memory and bandwidth efficient, as the agents can optimise their vocabulary size and word length to their specific task, providing an advantage over a general communication protocol \citep{rita_lazimpa_2020}. This approach also allows for inherent vocabulary alignment \citep{chocron_vocabulary_2020}, even allowing for alignment between agents which were not trained together through population approaches \citep{rita_role_2022,michel_revisiting_2022,rita_language_2024}, enabling convergence to a common language across a population of agents.

Many aspects of emergent language have been explored \citep{lazaridou_emergent_2020,boldt_review_2024}, with a particular focus on improving communication efficiency \citep{rita_lazimpa_2020,chaabouni_anti-efficient_2019,kang_incorporating_2020}. \citet{kang_incorporating_2020} demonstrate how using the minimal deviation between subsequent time steps allows for more concise communication by reducing redundant information transfer. Investigation of the contextual information of the resulting language offers a further improvement in agent performance by using the time step similarity together with optimisation of the reconstruction of the speaker's state \citep{kang_incorporating_2020}. There is, however, no existing research investigating or reporting on the emergence of temporal referencing strategies, where agents could communicate about relationships between different time steps.

Such temporal references, together with the general characteristics of emergent languages, have the potential to enhance the agents' bandwidth efficiency and task performance in a variety of situations. Temporal references are  the only way that agents can coordinate in tasks that require temporal understanding. For example, in a turn‐based strategy game an agent might need to refer back to “the opponent’s move two turns ago” to  communicate past behaviour; in multi‐robot monitoring one drone could report “the sensor reading from three scans ago” to its partner; in multi-agent coordination, agents must be able to schedule synchronisation times (“Sync next at t=10”) to preserve bandwidth \citep{wang_multi-agent_2023}. The simplified environment used in our study provides a controlled and interpretable setting that allows us to systematically probe the specific factors driving the emergence of such temporal references. While this setting enables clearer insights into the underlying mechanisms, applying these findings to more complex, real-world scenarios, such as social deduction games \citep{brandizzi_rlupus_2021,lipinski_emergent_2022,kopparapu_hidden_2022}, may introduce additional challenges due to the richer dynamics and strategic complexity involved. Nonetheless, our work offers a foundational step toward understanding how temporal communication strategies could emerge in such environments. 

Temporal references would also allow agents to develop more efficient methods of communication by assigning shorter messages to events that happen more often. This is similar to Zipf's Law in human languages \citep{zipf_human_1949}, which states that the most commonly used words are the shortest. This strategy would be particularly effective when the distribution of observations would be non-uniform, which means that certain objects appear more often than others. Specialised messages, used only for temporal references, would then also become more frequent than others. From information theory, we know that (adaptive) Huffman coding \citep{huffman_method_1952,knuth_dynamic_1985,vitter_design_1987} can assign shorter bit sequences to more frequent messages, thereby compressing them more efficiently than less common messages. Consequently, the incorporation of temporal references can enhance the efficiency of transmitting emergent language, optimising communication.

Our contribution lies in examining when temporal references emerge between agents. In this work we perform a focused study of three potential factors of temporal references \--- environmental pressures, external losses, and architectural biases \--- so as to pinpoint their individual effects without conflating other factors. The agents are trained in both the regular referential game \citep{lazaridou_multi-agent_2017} and on an environment which encourages the development of temporal references through embedded environmental pressures (\cref{ssec: env}). The effect of an external pressure to develop temporal referencing is explored via an additional loss applied to the agents (\cref{sec: experiments}). Three types of architecture are evaluated (\cref{sec: architectures}). The baseline  \basen (\cref{ssec: base}) agent, based on the commonly used EGG agents \citep{kharitonov_egg_2019}, provides us with a reference performance for both the emergence of temporal references, and performance in an environment. This baseline is compared to a \temporaln (\cref{ssec: temporal}) agent, which features a sequentially batched LSTM, instead of the parallel batching used in EGG, which allows the agents to build an understanding of the target sequence. Additionally, the \temporalrn (\cref{ssec: temporalr}) agent combines the information from the sequential LSTM and the parallel batched LSTM from the \basen agent. This allows it to process information about the objects, without needing to focus on their order in the sequence at the same time.

\section{Referential Game Environments}\label{sec: trg}

\begin{figure}
    \centering
    \begin{subfigure}{0.48\textwidth}
        \centering
        \includegraphics[width=0.9\textwidth]{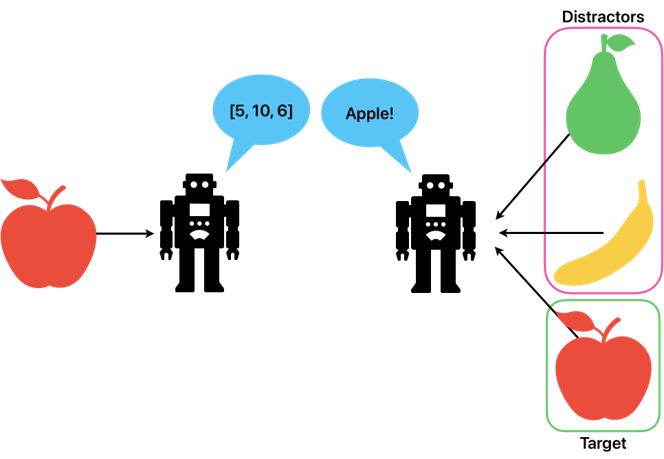}
        \caption{Referential Game}
        \Description[A visual representation of an emergent communication referential game, with two agents.]{Two agents in a referential game setting. The sender on the left agent observes a red apple, and emits a message "[5,10,6]". The receiver agent on the right observes a red apple, a yellow banana and a green pear, and correctly guesses the target object, by emitting the prediction for apple.}
        \label{fig: rg}
    \end{subfigure}
    \hfill
    \begin{subfigure}{0.48\textwidth}
        \centering
        \includegraphics[width=0.9\textwidth]{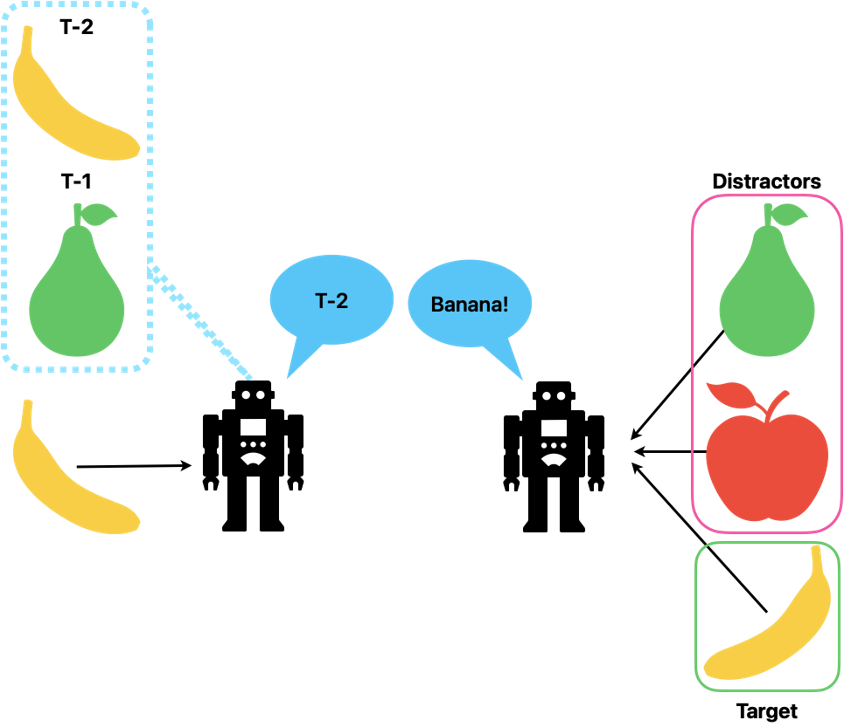}
        \caption{Temporal Referential Game}
        \Description[A visual representation of an emergent communication temporal referential game, with two agents.]{Two agents in a temporal referential game setting. The sender on the left agent observes a yellow banana, together with the previously seen objects which are a green pear, and an another yellow banana, at time steps T-1 and T-2 respectively. The sender emits a message "T-2". The receiver agent on the right observes a red apple, a yellow banana and a green pear, and correctly guesses the target object of yellow banana, by emitting the prediction for yellow banana, based on the information of the object being the same as two time steps ago.}
        \label{fig: trg}
    \end{subfigure}
    \caption{Structure of the referential game and temporal referential game.}
    \Description[A comparison of the temporal and regular referential games.]{The referential game visualisation is on the left, with the temporal referential game on the right.}
    \label{fig: games}
\end{figure}

Referential games \citep{lewis_convention_1969,lazaridou_emergence_2018} provide the most commonly used framework to study emergent communication, employing two agents: a sender and a receiver \citep{kharitonov_egg_2019}. The sender begins the game by observing a target object, and then generates a message. This message is passed to the receiver, along with the target object and a number of distractor objects. The receiver's task is to discern the target object from among the objects it observes, using the information contained in the message it receives. This exchange is repeated every episode. The referential game is presented in \cref{fig: rg}. For comprehensive reviews of these environments, we refer to \citet{lazaridou_emergent_2020} or \citet{brandizzi_toward_2023}.

In our referential games, we use attribute-value vectors to represent objects, enabling us to isolate and control factors that might impact agent performance. This approach \citep{kharitonov_egg_2019,chaabouni_compositionality_2020,ueda_categorial_2022} serves as an effective test-bed for investigating temporal properties of emergent language.

\subsection{Definitions}

In our referential games, agents identify objects from an \textit{object space} $V$, represented as attribute-value vectors $\bm{x} \in V$. The \textit{value space} $S = \{0,1,2 \ldots N_{val}\}$ defines possible variations for each attribute, where $N_{val}$ is the \textit{number of values}. The \textit{object space} is defined as $V = S_1\times\cdots\times S_N = \{(a_1, \ldots, a_{N_{att}}) \mid a_i \in S_i \ \text{for every} \ i \in \{1, \ldots, N_{att}\} \}$, where $N_{att}$ is the \textit{number of attributes}.

For intuition, consider an object as an abstraction of an image of a circle. Attributes might include line style (dashed/solid), line colour, and background colour. Values represent variations of these attributes (e.g., blue, red, or black for colour). A blue solid-line circle on a red background could be represented as [blue, solid, red] or as integers $[2,1,3]$.

The characters available to the agents (\ie \textit{the symbol space}) is $\omega = \{0,1,2 \ldots N_{vocab} - 1\}$ where $N_{vocab}$ is the \textit{vocabulary size}. The \textit{message space}, or the space that all messages must belong to, is defined as $\xi = \omega_1\times\cdots\times \omega_L = \{(c_1, \ldots, c_L) \mid c_i \in \omega_i \ \text{for every} \ i \in \{1, \ldots, L\} \}$, where $L$ is the maximum message length. The message space $\xi$ represents all possible symbol combinations that could constitute a message.

The agents' language maps objects in $V$ to messages in $\xi$. The exchange history is a sequence $\tau = \{(\bm{m}_n,\bm{x}_n)\}_{n\in\{1,...,t\}}$ such that $\forall n,  \bm{m}_n \in \xi \land \bm{x}_n \in V$, with $t$ representing the episode of the last exchange.

\subsection{Temporal Logic}\label{ssec: methods tl}

We use temporal logic to formally define our environment's behaviour and analyse agent communication. Specifically, we employ a form of \acrfull{ltl} \citep{pnueli_temporal_1977} called \acrfull{pltl} \citep{lichtenstein_glory_1985}.

While \acrshort{ltl} focuses on future-present connections using operators like "next" $\medcirc$, \acrshort{pltl} extends this to include past relationships through operators like "previously" $\ominus$. The "previously" operator must satisfy \cref{eqn: defs} \citep{maler_checking_2008}, where $\sigma$ refers to an agent's message at time $t$, and $\phi$ signifies the object seen by the agent.

\begin{equation}\label{eqn: defs}
    \begin{aligned}
        (\sigma, t)\models \ominus\phi &\leftrightarrow (\sigma, t-1) \models \phi
    \end{aligned}
\end{equation}

We use the shorthand notation $\ominus^{n}$ to indicate the $\ominus$ operator applied $n$ episodes back (\eg $\ominus^4 \phi \leftrightarrow \ominus\ominus\ominus\ominus \phi$).

\subsection{Temporal Referential Games}\label{ssec: env}

In our temporal referential game, the sender might observe either a new random object or a previously seen object from earlier episodes. For example, in a sequence of episodes, the objects shown to the sender might be $[\bm{a}, \bm{b}, \bm{a}, \bm{c}, \bm{b}]$, where $\bm{a}$ and $\bm{b}$ repeat from previous episodes. This temporal dimension introduces the possibility for agents to develop communication that references objects across time, potentially using more efficient representations than describing each object from scratch.

This temporal version of the referential games \citep{lewis_convention_1969,lazaridou_multi-agent_2017} is based on the ``previously'' ($\ominus$) \acrshort{pltl} operator.\footnote{Code is available at \url{https://github.com/olipinski/TRG}} At every game step $s_t$, the sender agent is presented with an input object vector $\bm{x}$ generated by the function $X(t, c, h_v)$, with a random \textit{chance} parameter $c$, the \textit{previous horizon} value $h_v$, and the current episode $t$.

\begin{equation}\label{eqn: trgprevious}
    X(t,c,h_v) = \begin{cases}
    \bm{x} & c=0\\
    \ominus^{h_v} \bm{x} = \tau_{t-{h_v}} & c=1
    \end{cases}
\end{equation}

The \textit{previous horizon} value is uniformly sampled, taking the value of any integer in the range $[1,h]$, where $h$ is the \textit{previous horizon} hyperparameter, to allow agents to develop temporal references of varying temporal horizons. The function $X(t, c, h_v)$ selects a target object to be presented to the sender using \cref{eqn: trgprevious}, either generating a new random target object or using the old target object. This choice is facilitated using the \textit{chance} parameter $c$, which is sampled from a Bernoulli distribution, parametrised by the repetition chance parameter $p$, \eg $p=0.5$. If $c=1$ a previous target object is used, and if  $c=0$ a new target object is generated. Both $c$ and $h_v$ are sampled every time a target object is generated.

For example, consider an episode at $t = 4$ with the sampled parameters $c = 1$ and $h_v = 2$. Suppose the agent has observed the following targets: $[\bm{j}, \bm{k}, \bm{l}]$. Given that $c = 1$, further to \cref{eqn: trgprevious}, the $\ominus^2$ ($\ominus^{h_v}$) target is chosen. The target sequence becomes $[\bm{j},\bm{k},\bm{l},\bm{k}]$, with the target $\bm{k}$ being repeated, as it was the second to last target. Now suppose that $c$ was sampled to be $c = 0$ instead. Further to \cref{eqn: trgprevious}, a random target $\bm{a}$ is generated, and the target sequence becomes $[\bm{j},\bm{k},\bm{l},\bm{a}]$.

This behaviour describes the environment \textit{TRG}, which represents the base variant of temporal referential games, where targets are randomly generated with a $p$ chance of repetition of a target from the previous horizon $[1,h]$.  We additionally implemented five more environment variants:

\begin{description}[noitemsep]
    \item[TRG]: Base temporal referential game with a chance $p$ of object repetition.
    \item[TRG Hard]: Same as TRG but with targets that differ from distractors in only one, randomly chosen attribute.
    \item[RG]: Classic referential game.
    \item[RG Hard]: Classic referential game with targets that differ from distractors in only one, randomly chosen attribute.
    \item[Always Same]: A subset\footnote{A subset is used as the target space grows exponentially with the number of attributes and values.} of all possible targets repeated sequentially, used to verify consistent temporal messaging.
    \item[Never Same]: No target repetition, used to verify if temporal messages are used only in temporal contexts, and a as sanity check.
\end{description}

In all datasets where repetition may occur, only the target object is repeated, while the distractor objects are randomly generated for each object set. Sample inputs and expected outputs for all environments are provided in \cref{apx: test environments}.

Since RG, RG Hard, and Never Same environments have minimal target repetitions (\cf \cref{apx: dataset details}, \cref{fig: dataset repeats}), we do not expect temporal references to emerge in these settings. We use these environments to validate our results, and to provide a baseline performance on the environments usually used in emergent communication research \citep{boldt_review_2024}.

\section{Architectures}\label{sec: architectures}

In emergent communication, both sender and receiver agents are typically built around a single recurrent neural network \citep{kharitonov_egg_2019}. In this paper, these standard agent architectures, based on an LSTM \citep{hochreiter_long_1997}, are compared to temporal LSTM architectures, which use a different batching strategy. The \basen agent, used as a baseline, is the commonly used LSTM-based agent \citep{hochreiter_long_1997,kharitonov_egg_2019}. Our two novel architectures, the \temporaln and \temporalrn agents, feature a sequentially batched LSTM. This additional module allows the agents to gather information about the sequence of objects itself, instead of the regular LSTM which processes all objects in parallel. While the \temporaln agents use just the sequential LSTM to process their input, the \temporalrn  agents combine both the \basen and \temporaln approaches, combining the outputs of a regular LSTM, together with the sequentially batched LSTM. This allows the \temporalrn  agents to process the information that may be present in the objects themselves, as well as the order of their appearances, without needing to store this information in a single LSTM hidden state.

While many architectures may work for processing temporal information within datasets, we opt for a distinct batching strategy for two key reasons. Firstly, it entails the least modification to the \basen agent design, facilitating direct comparisons between the two architectures and enabling straightforward application of our architectural modifications to other contexts. Secondly, it offers a straightforward framework for examining the emergence of temporal references. Although our experiments initially involved more complex architectures, including attention-based agents, we observed no significant differences in any metrics from those of LSTM-based networks. Consequently, our focus in this study remains on LSTM-based agents.

\subsection{\basen Agent}\label{ssec: base}

In common with other approaches \citep{kharitonov_egg_2019,chaabouni_anti-efficient_2019,auersperger_defending_2022}, each of the \basen sender and receiver agents are constructed around a single LSTM \citep{hochreiter_long_1997}. First, the sender's LSTM  receives each target and distractor set individually, processing each object separately. The result is the initial hidden state for the message generation LSTM. For message generation, the same method is followed as used in previous work \citep{kharitonov_egg_2019}, and messages are generated character by character, using the Gumbel-Softmax trick \citep{jang_categorical_2017}. These messages are then passed to the receiver. The receiver's architecture contains an object embedding linear layer and a message processing LSTM, similar to the most commonly used architecture  \citep{kharitonov_egg_2019}. In the receiver agent, a hidden state is computed for each message by the LSTM. This output is combined with the output of the object embedding linear layer to create the referential game object prediction. The architecture diagram for the \basen agent is available in \cref{apx: arch details}, \cref{fig: arch-base-lstm}.

\subsection{\temporaln Agent}\label{ssec: temporal}

\begin{figure}
    \centering
    \begin{subfigure}{0.48\textwidth}
        \centering
        \includegraphics[width=0.9\textwidth]{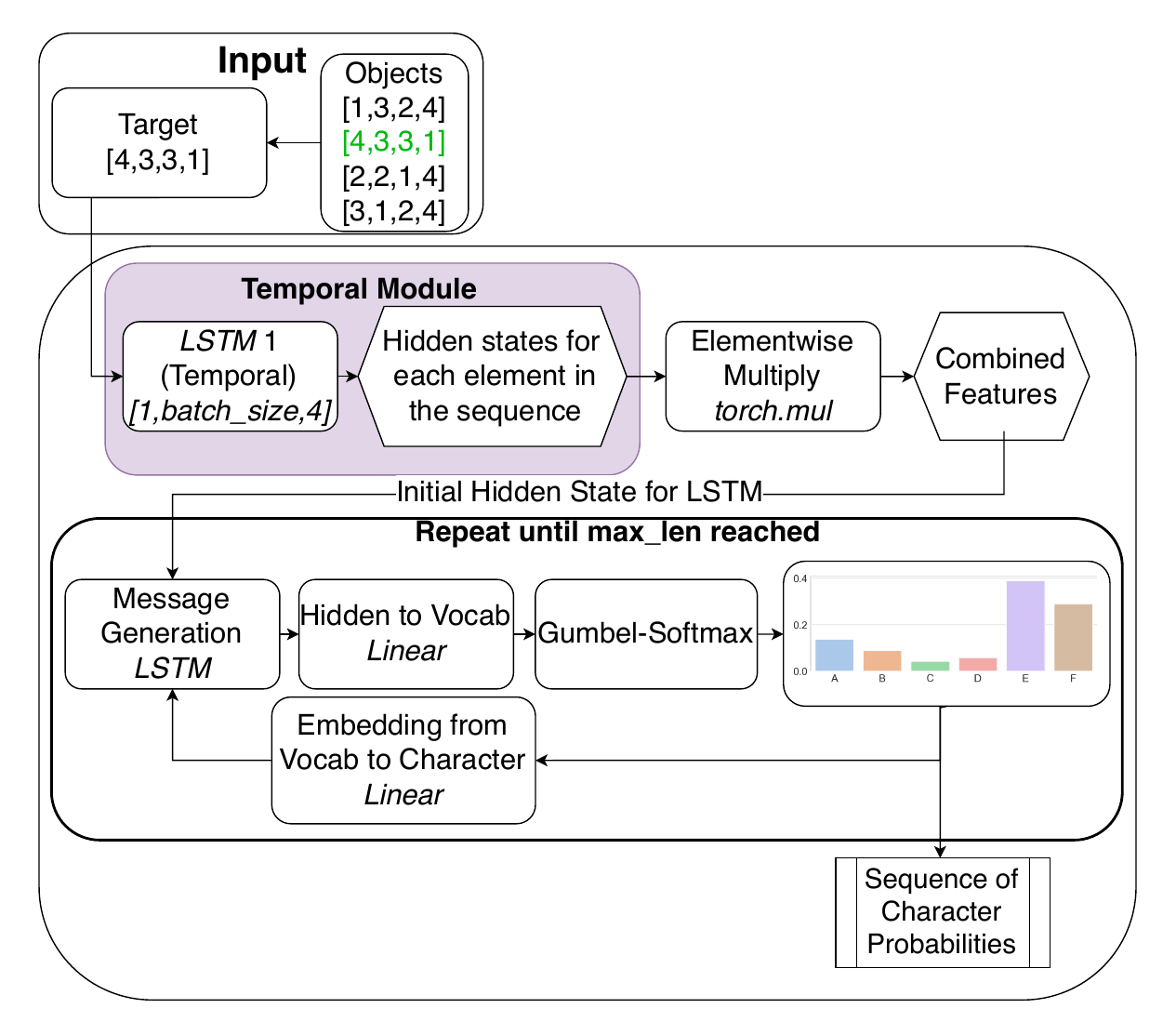}
        \caption{The \temporaln LSTM sender architecture.}
        \label{fig: arch-sender-t-lstm}
    \end{subfigure}
    \hfill
    \begin{subfigure}{0.48\textwidth}
        \centering
        \includegraphics[width=0.9\textwidth]{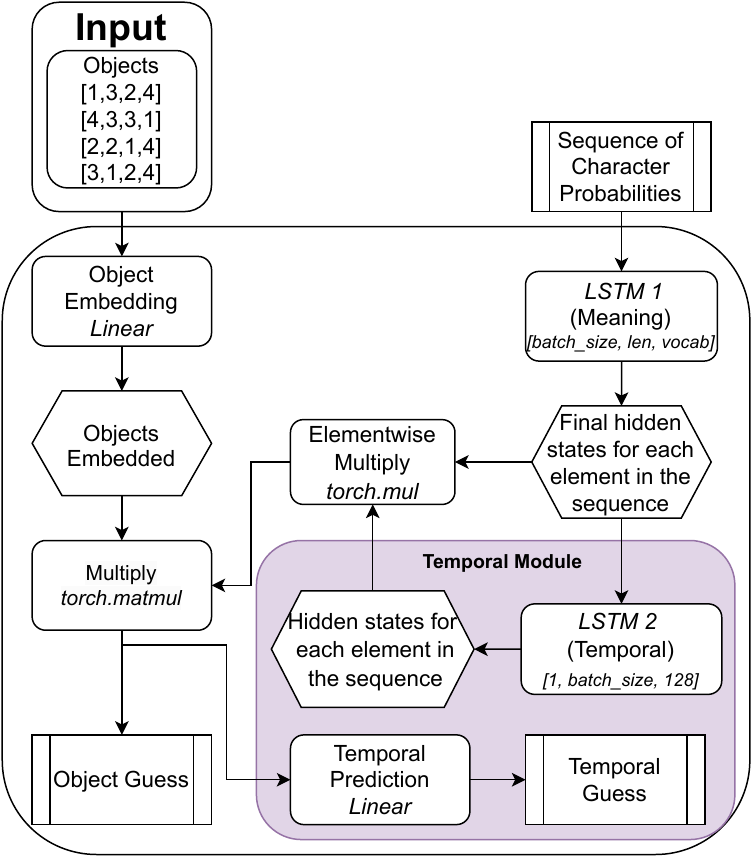}
        \caption{The \temporaln LSTM receiver architecture.}
        \label{fig: arch-receiver-t-lstm}
    \end{subfigure}
    \caption{The \temporaln LSTM sender and receiver architectures, with the temporal modules highlighted in purple.}
    \label{fig: arch-t-lstm}
\end{figure}

The \temporaln agent extends the \basen architecture by introducing a sequential LSTM module in both the sender and receiver networks (\cref{fig: arch-t-lstm}). This module enables the agents to capture temporal relationships across objects by processing sequences rather than treating each object as independent, similar to the sequential learning of language in humans \citep{christiansen_language_2003}. The module's aim is to allow the agents to develop a temporally grounded understanding of the input, supporting the emergence of temporal references.

\begin{figure}[ht]
    \centering
    \begin{subfigure}{0.48\textwidth}
        \centering
        \includegraphics[width=0.8\textwidth]{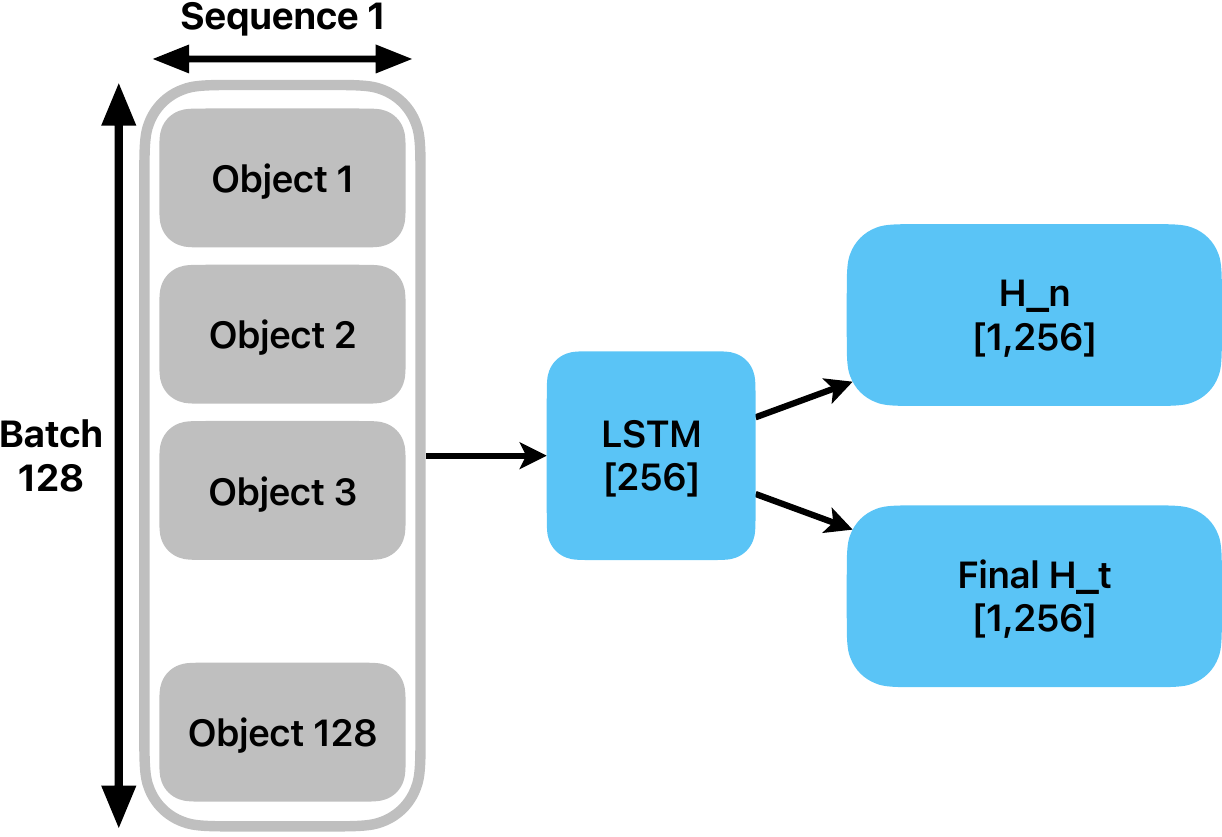}
        \caption{Regular batching of an LSTM.}
        \label{fig: reg-batch}
    \end{subfigure}
    \hfill
    \begin{subfigure}{0.48\textwidth}
        \centering
        \includegraphics[width=0.85\textwidth]{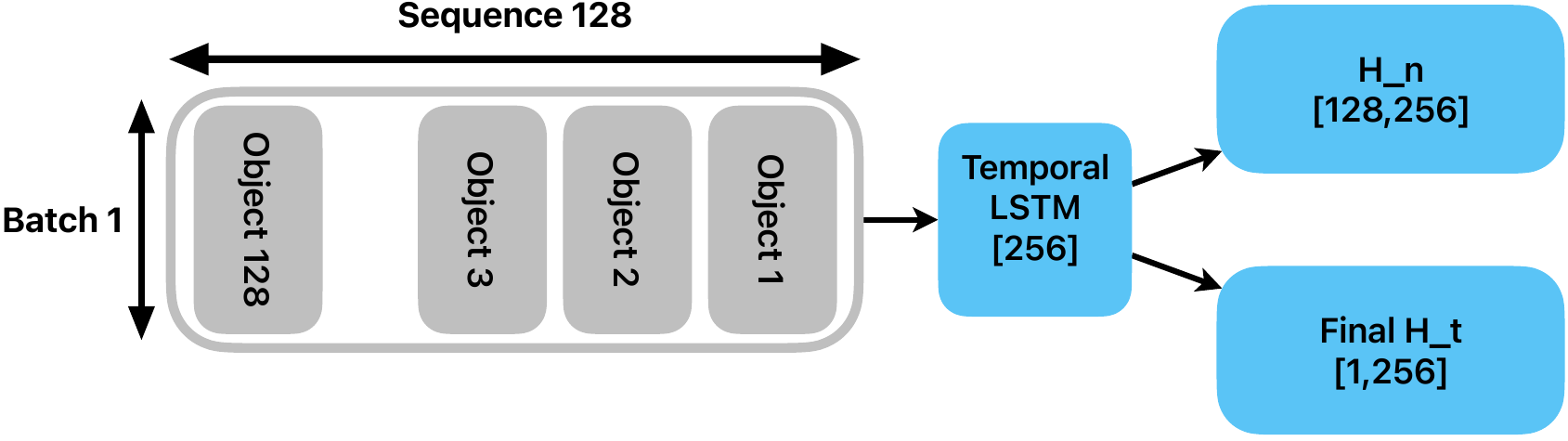}
        \caption{Temporal batching of an LSTM.}
        \label{fig: t-batch}
    \end{subfigure}
    \caption{Examples of regular and temporal batching strategies.}
    \label{fig: batching}
\end{figure}

In standard LSTM architectures \citep{kharitonov_egg_2019}, each object is processed independently in parallel. This corresponds to input batches of shape, \eg $[128, 1, 6]$, where $128$ is the batch size and each object has $6$ attributes. In contrast, the \temporaln agent uses a sequential batching strategy, changing the batch shape to $[1, 128, 6]$, or a single sequence of $128$ objects. This allows the LSTM to process the objects in temporal order and develop a representation of the sequence as a whole. Visualisations of both approaches are shown in \cref{fig: batching}.

In the sender (\cref{fig: arch-sender-t-lstm}), the output of the sequential LSTM provides the initial hidden state for message generation. The message generation procedure follows the standard approach, where messages are produced character by character using the Gumbel-Softmax trick \citep{jang_categorical_2017}. These messages are then passed to the receiver.

The receiver (\cref{fig: arch-receiver-t-lstm}) mirrors the sender's design. It first processes the message using a standard LSTM to obtain a set of hidden states, then feeds these into a sequential LSTM. This additional LSTM captures temporal dependencies in the message structure and supports the identification of recurring objects. The resulting temporal representation is combined with object embeddings for referential prediction.

To further promote temporal reasoning, the receiver may include a temporal prediction module. This module computes a label representing how far back (in number of episodes) an object was last seen, up to a fixed previous horizon $h$. This label is predicted using a single linear layer. For instance, if an object last occurred $5$ episodes ago and $h = 8$, the label is $5$ since $5 \le h$. If instead $h = 4$, the label defaults to $0$, as the previous occurrence lies beyond the horizon.

Training includes an auxiliary loss term for this prediction, combined with the standard referential game loss. The total loss function is defined as $L_t = L_{rg} + L_{tp}$, where $L_{rg}$ is the referential loss (cross-entropy between the receiver's guess and the sender's target), and $L_{tp}$ is the temporal prediction loss (cross-entropy between the predicted and true temporal labels). This auxiliary objective helps focus learning on temporal relationships, encouraging the development of temporally grounded communication. Further analysis of its effect is presented in \cref{sec: experiments}.

\subsection{\temporalrn Agent}\label{ssec: temporalr}

The \temporalrn  agent combines the \basen and the \temporaln architectures. The sequential LSTM from the \temporaln agent is added to the \basen architecture, merging both the sequential understanding from the temporal module, with the parallel understanding of the target objects from the regularly batched LSTM. The hidden states of both LSTMs are combined through an element-wise multiplication and fed into the message generation LSTM. The receiver agent is the same as the \temporaln receiver, and includes the temporal prediction module. The architecture diagram for the \temporalrn agent is available in \cref{apx: arch details}, \cref{fig: arch-tr-lstm}.

\section{Metrics}\label{sec: metrics}

\subsection{Temporality Metric}\label{ssec: temporlity metric}

To evaluate the development of temporal references, we propose a new metric, $M_{\ominus^n}$, which measures how often a given message was used as the ``previous'' operator. Given an exchange history (the sequence of objects shown to the sender and messages sent to the receiver), $\tau$, it checks when an object has been repeated within a given horizon $h_v$, and records the corresponding message sent to describe that object. The objective of the $M_{\ominus^n}(\bm{m})$ metric is to measure if the message can give reference to a previous episode; \eg if a message is used similarly to the sentence ``The car I can see is the same colour as the one mentioned two sentences ago''.

Let $C_{\bm{m}}{\ominus^n}$ count the times the message $\bm{m}$ has been sent together with a repeated object for $h_v=n$:
\begin{equation}
    C_{\bm{m}{}\ominus^n} = \sum_{j=1}^{t} \mathbb{I}(\bm{m}_j = \bm{m} \land \text{objectSame}(\bm{x}_j, n))
\end{equation}
where $\mathbb{I}(\cdot)$ is an indicator function that returns $1$ if the condition is true and $0$ otherwise, and $\text{objectSame}(\bm{x}_j, n)$ is a function that evaluates to true if the object $\bm{x}_j$ is the same as the object $n$ episodes ago.

Let $\mathcal{T}_{\bm{m}}$ denote the total count of times the message $\bm{m}$ has been used:
\begin{equation}
    \mathcal{T}_{\bm{m}} = \sum_{j=1}^{t} \mathbb{I}(\bm{m}_j = \bm{m})
\end{equation}
where $\mathbb{I}(\cdot)$ is an indicator function selecting the message $\bm{m}$ in the exchange history $\tau$.

The percentage of previous messages that are the same as $\bm{m}$ can then be calculated using $M_{\ominus^n}(\bm{m})$:
\begin{equation}
    M_{\ominus^n}(\bm{m}) = \frac{C_{\bm{m}}{\ominus^n}}{\mathcal{T}_{\bm{m}}} \times 100
\end{equation}

The $M_{\ominus^n}$ metric is sensitive to repeated object appearances due to dataset structure, which may lead to false positives in environments with high repetition probability. If a language does not have temporal references and uses a given message to describe an object consistently, this message will be repeated every time this object appears. This means that for every repetition object, the message could be considered to indicate a \textit{previous} episode, whereas, in reality, it is just a description of the object. For example, if a dataset contains 75\% repetitions, on average, each message will be used as an accidental $\ominus^n$ 75\% of the time. 

Therefore, in \cref{sec: experiments}, the emergence of temporal references is defined as the appearance of messages that reach 100\% on the $M_{\ominus^n}$ metric. This strict threshold helps avoid ambiguity: a message used 30\% or 70\% of the time in a temporal context might reflect genuine referencing, but it could just as easily result from consistent object labelling or a high repetition rate in the dataset. Simply comparing against the base repetition rate in the dataset is also not enough; while the chance of repetition can be estimated (and is close to 0\% in RG environments, as shown in \cref{apx: dataset details}), this does not help distinguish between intentional temporal reference and accidental reuse of a message. Without additional heuristics or manual inspection, it would be difficult to differentiate the false positives from the true positives. By focusing only on messages that appear exclusively in reference to previous episodes, \ie those with a perfect 100\% score, we ensure that the signal is both interpretable and robust.

Importantly, even if only a single message reaches this 100\% threshold, we still consider it valid evidence for the emergence of temporal reference. The only scenario in which a message could reach 100\% without carrying temporal meaning is a degenerate one \--- a dataset where the same object is repeated every time. Otherwise, consistent and exclusive use of a single message in repeated contexts indicates that the agents have developed a specific, maximally efficient marker for temporal reference; something highly desirable, as such a message is both highly compressible and interpretable.

\paragraph{Example} Consider an object sequence $[\bm{x},\bm{y},\bm{z},\bm{y},\bm{y},\bm{y},\bm{x}]$ with three possible corresponding message sequences:
\begin{enumerate}
    \item $[\bm{m}_1,\bm{m}_2,\bm{m}_3,\bm{m}_2,\bm{m}_4,\bm{m}_4,\bm{m}_1]$
    \item $[\bm{m}_1,\bm{m}_2,\bm{m}_3,\bm{m}_4,\bm{m}_4,\bm{m}_4,\bm{m}_1]$
    \item $[\bm{m}_1,\bm{m}_2,\bm{m}_3,\bm{m}_2,\bm{m}_2,\bm{m}_2,\bm{m}_1]$
\end{enumerate}

Assume horizon $n=1$ (i.e., $\ominus^1$). There are two repetitions in the sequence of objects: the second and third $\bm{y}$ following the sequence of $[\bm{x},\bm{y},\bm{z},\bm{y}]$.

In message sequence (1), $\bm{m}_4$ is used exactly for the second and third repetitions of $\bm{y}$ and never otherwise: $C_{\bm{m}_4}{\ominus^1} = 2$, $\mathcal{T}_{\bm{m}_4} = 2$, and $M_{\ominus^1}(\bm{m}_4) = 100\%$.

In message sequence (2), $\bm{m}_4$ is also used for the initial (non-repeating) instance, giving $C_{\bm{m}_4}{\ominus^1} = 2$, $\mathcal{T}_{\bm{m}_4} = 3$, and $M_{\ominus^1}(\bm{m}_4) = 66\%$.

In message sequence (3), $\bm{m}_2$ is used both for the initial and repeated instances of $\bm{y}$: $C_{\bm{m}_2}{\ominus^1} = 2$, $\mathcal{T}_{\bm{m}_2} = 4$, and $M_{\ominus^1}(\bm{m}_2) = 50\%$.

This example further illustrates why only messages scoring $100\%$ are considered to signify a temporal reference. 

\subsection{Correctness Metric}\label{ssec: metrics correctness}

\textit{Correctness} quantifies how often the receiver agent correctly identifies the target object upon receiving a given message. For a message $\bm{m}$ used $N$ times, if the receiver correctly guesses the target in $K$ of those cases, the correctness score is:
\begin{equation}
    \text{Correctness}(\bm{m}) = \frac{K}{N} \times 100\%
\end{equation}
This metric captures mutual intelligibility between sender and receiver: a high score indicates consistent, shared interpretation of the message.

\subsection{Compositionality Metrics}\label{ssec: compositionality metrics}

The emergent languages are also analysed in terms of their compositionality scores, using the topographic similarity metric \citep{brighton_understanding_2006}, commonly employed in emergent communication. Topographic similarity measures the Spearman Rank correlation \citep{spearman_proof_1904} between the distances of messages and objects in their respective spaces. For example, a message describing a ``blue circle'' should be closer to ``blue triangle'', than to ``red square'', given the language is compositional. 

The languages are also evaluated using metrics which account for languages where the symbols themselves carry all the information: \textit{posdis}, which use the positional information of individual characters, and \textit{bosdis}, for permutation invariant languages,  \citep{chaabouni_compositionality_2020}. Posdis intuitively measures if, for example, the first symbol always refers to a property of the object, such as in the English phrases ``blue circle'' or ``red square'', versus ``circle blue'' or ``square red''. Bosdis measures whether a symbol carries all the information independent of the position of this symbol, such as in the English conjunctions example from \citet{chaabouni_compositionality_2020},  ``dogs and cats'' and ``cats and dogs'', where both constructions are valid and convey the same information.

\section{Experiments}\label{sec: experiments}

To study the impact of the architecture, as well as the external and internal pressures of the agents, we propose three hypotheses to guide our experiments:

\begin{description}[noitemsep]
    \item[Hypothesis 1 (H1)] All agents can develop some form of temporal references, with agents that include the temporal prediction loss more likely to do so.
    \item[Hypothesis 2 (H2)] Temporal references will increase the agent's performance in environments which include temporal relationships.
    \item[Hypothesis 3 (H3)] No temporal references will be detected with the $M_{\ominus^n}$ metric in environments where there are no temporal relationships.
\end{description}

Hypothesis 1 is investigated by using multiple agent types and applying a temporal prediction loss. Hypothesis 2 is analysed by using two environments, TRG and TRG Hard, where temporal relationships are explicitly introduced. We then compare the performance of agents which develop temporal references, to those which do not to determine the impact of temporal references on the task accuracy. We include all evaluation environments, including those where there are no target repetitions when evaluating the $M_{\ominus^n}$ to validate Hypothesis 3, which acts as a sanity check.

To test our hypotheses, the following architectures are trained and evaluated:
\begin{description}[noitemsep]
    \item[\basen] The same as the EGG \citep{kharitonov_egg_2019} agents, used as a baseline for comparisons (\cref{ssec: base});
    \item[\temporaln] The base learner with the sequential LSTM \textit{instead} of the regularly batched LSTM (\cref{ssec: temporal}); and
    \item[\temporalrn] The base learner with the regularly batched LSTM \textit{and} the sequential LSTM (\cref{ssec: temporalr})
\end{description}

Each agent type is additionally trained with and without the temporal prediction loss. The agents that include the temporal prediction loss have an explicit reward to develop temporal understanding. There is no additional pressure to develop temporal references for agents that do not include the temporal prediction loss, except for the possibility of increased performance on the referential task. The aim of the additional loss is to investigate if it would aid in the temporal reference emergence, or if it improves agent performance. We show all the evaluated agent configurations in \cref{apx: experimental combinations}.

All agent configurations were trained for the same number of epochs and on the same environments during each run. Agent pairs are trained in either the RG or TRG environment. Evaluation is performed after the training has finished in all environments \cref{ssec: env}. All target objects are uniformly sampled from the object space $V$. Each possible configuration was run ten times, with randomised seeds for both the agents and the datasets. We vary the target repetition chance, $p \in \{0.25, 0.5, 0.75\}$, in the training datasets to examine how repetition frequency affects the development of temporal references. \cref{apx: training details} provides further details on all training hyperparameters. 

\subsection{Evaluation and Results}\label{ssec: temporality analysis}

The metric value distribution of each network type is analysed, using the Kruskal-Wallis H-test \citep{kruskal_use_1952}, as scores are not guaranteed to be distributed normally. Conover-Iman \citep{conover_multiple-comparisons_1979} post-hoc analysis is performed, with Holm-Bonferroni \citep{holm_simple_1979} corrections applied, to verify which of the different network types differ significantly from each other. Using these methods, all results reported in the following sections have been verified to be statistically significant, with $p<0.05$. The detailed results of the significance analyses are shown in the supplemental material. 

\paragraph{Task Accuracy}

All agents achieve high task accuracy, with all achieving over 95\% in the Referential Games (RG) environment. Both Hard variants (\ie RG Hard and TRG Hard) present a challenge to the agents. All agents perform significantly worse in these two evaluation environments, achieving approximately 72\% accuracy on average for the RG Hard environment, and 85\% accuracy for the TRG Hard environment. We observe no increase in the task accuracy for the agents which develop temporal references, proving H2 incorrect. We provide a discussion of the possible reasons behind this in \cref{sec: discussion}. The detailed accuracy distributions are provided in \cref{apx: accuracy}.

\paragraph{Temporal Analyses}

In line with H3, the values of $M_{\ominus^n}$ for the Never Same, RG, and RG Hard environments consistently register at $0\%$. These results reduce the likelihood of significant issues with the $M_{\ominus^n}$ metric, given that the probability of target repetition in these environments is near zero. Since the results remain constant and at $0\%$ for these environments, they are omitted from the subsequent sections for brevity. Detailed results for these environments are available in both \cref{apx: accuracy,apx: topsim}, and supplemental material.

\cref{tab: m4 values} illustrates the $M_{\ominus^4}$ metric values, referring to an observation four messages in the past, of all agent types over the evaluation environments (\cf \cref{sec: experiments,sec: metrics}), where $M_{\ominus^4} \geq 0\text{\%}$.\footnote{The value of $4$ is chosen arbitrarily, to lie in the middle of the explored range of $h$. More detailed analysis from $h_v=1$ to $h_v=8$ is provided in the supplemental material} \cref{tab: m4 values} indicates that the temporally focused processing of the input data makes the agents predetermined to develop temporal references. Only the sequential LSTM (\temporaln and \temporalrn) agents, are capable of producing temporal references. Temporal references emerge in these networks, regardless of the training dataset. Even in a regular environment, without additional pressures, temporal references are advantageous. No messages in the \basen architectures are used $100\%$ of the time for $\ominus^4$, irrespective of the dataset they have been trained on. The temporal prediction loss is not enough, and the ability to process observations temporally is the key factor to the emergence of temporal references. These results partially confirm H1, indicating that all agents capable of explicitly processing temporal relationships develop temporal references and that no additional pressures are required. While we expected that the additional temporal prediction loss would improve the development of temporal references, this analysis indicates that it is not sufficient, nor necessary.

Additionally, messages that are used for $\ominus^4$ have a high chance of being correct, with most averaging above 90\% correctness (\cref{ssec: metrics correctness}). A detailed overview of the distribution of message correctness is provided in the supplemental material.

\begin{table}[ht]
    \centering
    \caption{Maximum value of the $M_{\ominus^4}$ metric for each network/loss/training environment combination. ``AS'' is Always Same, ``RG'' is the regular Reference Game environment, ``TRG'' is the Temporal Reference Game, and ``TRG Hard'' is TRG with the target differing in a single attribute with respect to the distractors.}
    \begin{tabular}{lllllllll}
         Network & Loss & Training Env & AS & TRG & TRG Hard  \\
         \toprule
         Base & Reg & RG & 60\% & 85\% & 85\% \\
         Base & Reg & TRG & 60\% & 85\% & 85\% \\
         Base & Reg+T & RG & 60\% & 85\% & 85\% \\
         Base & Reg+T & TRG & 60\% & 85\% & 85\% \\
         \midrule
         Temporal & Reg & RG & \textbf{100\%} & \textbf{100\%} & \textbf{100\%} \\
         Temporal & Reg & TRG & \textbf{100\%} & \textbf{100\%} & \textbf{100\%} \\
         Temporal & Reg+T & RG & \textbf{100\%} & \textbf{100\%} & \textbf{100\%} \\
         Temporal & Reg+T & TRG & \textbf{100\%} & \textbf{100\%} & \textbf{100\%} \\
         \midrule
         TemporalR & Reg & RG & \textbf{100\%} & \textbf{100\%} & \textbf{100\%} \\
         TemporalR & Reg & TRG & \textbf{100\%} & \textbf{100\%} & \textbf{100\%} \\
         TemporalR & Reg+T & RG & \textbf{100\%} & \textbf{100\%} & \textbf{100\%} \\
         TemporalR & Reg+T & TRG & \textbf{100\%} & \textbf{100\%} & \textbf{100\%} \\
    \end{tabular}
    \label{tab: m4 values}
\end{table}

Most messages are used only in the context of the current observations, with \temporaln and \temporalrn networks using a more specialised subset of messages to refer to the temporal relationships. Only \temporaln and \temporalrn variants develop messages that reach $100\%$ on the $M_{\ominus^4}$ metric. The distribution also suggests that these messages could be a more efficient way of describing objects, as the number of temporal messages is relatively small. Since only a small number of messages are needed for the temporal references, they could be used more frequently. This message specialisation, combined with a linguistic parsimony pressure \citep{rita_lazimpa_2020}, could lead to a more efficient way of describing an object: sending the object properties requires more bandwidth than sending only the time step the object last appeared. A detailed breakdown of the messages used is provided in the supplemental material.

The percentage of networks that develop temporal messaging is shown in \cref{tab: emergence count percentage}. The percentages shown are absolute values, calculated by taking the total number of runs and checking whether at least one message has reached $M_{\ominus^n} = 100\text{\%}$ for each run. That number of runs is divided by the total number of runs of the corresponding configuration to arrive at the quantities in \cref{tab: emergence count percentage}. 

The \temporaln and \temporalrn network variants reach over $96\%$ of runs that have converged to a strategy which uses at least one message as the $\ominus^n$ operator. In contrast, the \basen networks never achieve such a distinction. However, some runs have not converged to a temporal strategy in the case of the \temporalrn network. These instances account for only $3\%$ of the total number of runs, and the differences are not statistically significant from the \temporaln network, showing that the emergence of temporal references is still very likely, if somewhat dependent on the network initialisation. These results indicate that the ability to build a temporally focused representation of the input data is the deciding factor in the emergence of temporal references.

\begin{table}[h]
        \centering
        \caption{Percentage of networks that develop temporal messages.}
        \begin{tabular}{lll}
            Network Type & Loss Type & Percentage  \\
            \toprule
            Base & Regular loss & $0\%$ \\
            Base & Regular + Temporal loss & $0\%$ \\
            Temporal & Regular loss & $100\%$ \\
            Temporal & Regular + Temporal loss & $100\%$ \\
            TemporalR & Regular loss & $98.66\%$ \\
            TemporalR & Regular + Temporal loss & $97\%$ \\
        \end{tabular}
        \label{tab: emergence count percentage}
\end{table}

When the repetition chance $p$ increases, the percentage of messages that are used for $\ominus^n$ increases for all agent variants. We observe no other notable differences (\eg differing temporal strategies or emergence rates) between the different values of $p$; therefore, this is the only difference reported. On average, \basen networks demonstrate the same chance of using a message for $\ominus^n$ as the dataset repetition chance. This means that while the percentage increases, it is only due to the increase in the repetition chance. In contrast to the \basen networks, for \temporaln and \temporalrn networks, the average percentage does reach 100\%. This means that messages the agents designate for $\ominus^n$ are used more often than the repetition chance.

\paragraph{Language Analyses}

All agents create compositional languages with varying degrees of structure, which shows that learning to use temporal references does not negatively impact compositionality. All agents reach values between 0.1 and 0.2 (the higher, the more compositional the language is) on the topographic similarity metric \citep{brighton_understanding_2006,rita_emergent_2022}, where a score of 0.4 has been considered high in previous research \citep{rita_emergent_2022}. We provide a visual representation of the topographic similarity scores in \cref{apx: topsim}. These results indicate that temporal references have no negative effect on the languages' compositionality, showing that their emergence does not necessitate a trade-off in the possible generalisation ability of the emergent language \citep{auersperger_defending_2022}.

The differences between the \textit{posdis} and \textit{bosdis} distributions for each network type are not statistically signifiant. Therefore, the differences in the \textit{posdis} and \textit{bosdis} metrics across network types could be due to random fluctuations in the score distribution.

Analysing the development of temporal references, we observe that agents trained with a temporal module learn to use messages that refer consistently to prior object occurrences.

As an example, in one training run, using the \temporalrn configuration, a specific message, $\bm{m_t} = [25, 6, 9, 3, 2]$, consistently emerged as a temporal reference, or a $\ominus^1$ operator. During evaluation in the Always Same environment, this message was used exclusively in cases where the target object had appeared in previous observations. Notably, it was used across twelve distinct objects, but only after their initial appearance.

For each of these twelve objects, which were repeated ten times each during evaluation, this message was used in nine out of ten occurrences. A different message was used only when the object appeared for the first time. For instance, when the object $[4, 2, 3, 6, 5, 8, 8, 4]$ first appeared, the agents produced a different message, $[25, 6, 17, 9, 9]$. On subsequent occurrences, however, the temporal message $\bm{m_t}$ was used instead.

This consistent pattern demonstrates that the agents learn to generalise temporal references developed during training. Importantly, the same temporal message learned in the TRG environment was successfully transferred and reused in the Always Same environment, despite the object sets being different. This suggests that the agents do not rely on object identity alone but indeed use temporal references to generalise their behaviour across tasks.

\section{Discussion}\label{sec: discussion}
The results in \cref{ssec: temporality analysis} indicate that no explicit pressures are required for temporal messages to emerge, unlike increasing linguistic parsimony where additional losses are needed \citep{rita_lazimpa_2020,kalinowska_situated_2022}. We show that the incentives are already present in datasets that are \emph{not} altered to increase the number of repetitions occurring. Temporal references therefore emerge naturally, as long as the agents are able to build a temporal understanding of the data, such as with the sequential LSTM in the \temporaln and \temporalrn agents. This allows temporal references to emerge in any communication settings if a suitable architecture is used. This could provide greater bandwidth efficiency by allowing agents to use shorter messages for events that happen often, when combined with other linguistic parsimony approaches \citep{rita_lazimpa_2020,chaabouni_anti-efficient_2019}.

The emergence of temporal references only through architectural changes could also point towards additional insights in terms of modelling human language evolution using emergent communication \citep{galke_emergent_2022}. These architectural approaches to the emergence of temporal references could be viewed as analogous to sequential learning in natural language \citep{christiansen_language_2003}, as we learn to encode and represent elements in temporal sequences.

By focusing only on the three main factors for temporal reference emergence \--- architectural capacity, dataset repetition, and a temporal loss \--- we can confidently attribute emergence to the sequential understanding built by the temporal module. Future work can build on this foundation by exploring more complex features and aspects of temporal references in emergent communication. Such cases could include how often the agents use temporal references, and what combination of factors influence the agent's choices to use temporal references or to instead describe the object directly. 

\subsection{Accuracy}

As expected, both the RG Hard and the TRG Hard environments posed a challenge, presenting a significant accuracy drop. In the case of the \temporaln, \temporalrn agents, we hypothesised the emergence of temporal references to provide an advantage, increasing the agents' accuracy (\cref{sec: experiments}). While there indeed is a small increase in accuracy in the TRG Hard environment, it is not attributable to temporal references (\cf \cref{apx: accuracy}). This is because we observe the same increase for the \basen agents, which do not develop temporal references. We conclude that the increase is due to increased object repetition, making the task slightly easier.

A possible cause for the lack of accuracy increase for agents which develop temporal references might be the perceptual similarities between the highly similar distractor objects. This makes the task of discerning the difference between these objects becoming too difficult for the receiver. Alternatively, if the receiver struggled to correctly identify an object the first time it has observed it, the additional information that temporal references would offer would not be enough. The receiver would not know what the correct choice was for the previous timesteps.

Additionally, networks that have been trained with the temporal loss \textit{and} on the temporally focused dataset, perform slightly worse, by about $1\%$. The reason for the accuracy drop could lie in too much pressure on the temporal aspects of the dataset. Because of the additional loss, agents can increase their rewards by only focusing on creating temporal messages, without learning a general communication protocol. This then leads to overfitting the training dataset, where they can rely on both their mostly temporal language and their memory of the object sequences, instead of communicating about the object attributes. Consequently, we observe a decline in performance on the evaluation dataset. 

\subsection{Compositionality}

We hypothesise the effect of the temporal loss on the topographic similarity scores (we do not discuss the posdis and bosdis scores, as there are no statistically significant differences) is similar to that of the task accuracy. Since the agents focus on the temporal aspects of the task and dataset more, they develop less general languages, leading to lower compositionality scores. We do not believe the low compositionality scores are related to the simplicity of the dataset, being composed of integer vectors, since other works in similar settings achieve higher topographic similarity scores \citep{chaabouni_compositionality_2020,rita_emergent_2022}.

\section{Limitations}\label{sec: limitations}

All reported compositionality scores could be negatively affected by the presence of temporal references. Temporal messages can be compositional, but they would not refer to a specific object, and so topographic similarity would not be able to identify them correctly. Similarly, since posdis and bosdis also rely on the mappings between the messages and the dataset, instead of the temporal relationships captured by the temporal references, they could also be inaccurately lowered by their presence. Since temporal references, even if they were compositional, do not map directly to object properties in the dataset, they would be counted as non-compositional messages, therefore lowering the values of the evaluated metrics. This could be the reason for the lower values observed in our experiments when compared to previous research \citep{rita_emergent_2022}.

\section{Conclusion}\label{sec: conclusion}
Emergent communication has been studied extensively, considering many aspects of emergent languages, such as efficiency \citep{rita_lazimpa_2020,chaabouni_anti-efficient_2019}, compositionality \citep{auersperger_defending_2022}, generalisation \citep{chaabouni_compositionality_2020} and population dynamics \citep{chaabouni_emergent_2022, rita_role_2022, michel_revisiting_2022}. Yet, there has been no evidence of temporal relationships emerging. Discussing past observations is vital to communication, saving bandwidth by avoiding repeating information and allowing for easier experience sharing in multiagent communication.

This paper provides a first exploration of such emergent languages, including addressing the fundamental questions of when they could develop and what is required for their emergence. To achieve this, we confine our investigation to three main factors, enabling a targeted analysis of temporal reference emergence without potential confounding variables from more complex setups. We present a set of environments that are designed to facilitate investigation into how agents might create such references. We use the conventional agent architecture for emergent communication \citep{kharitonov_egg_2019} as a baseline and explore both temporal loss and alternative architectures that may endow agents with the ability to learn temporal relationships. We show that architectural change is necessary for temporal references to emerge, and demonstrate that temporal prediction loss is neither sufficient for their emergence, nor does it improve the emergent language. Our findings highlight the importance of analysing architectural designs in the study of multiagent communication, providing valuable insights for future research using environments where temporal relationships are important or into how other linguistic aspects might emerge.

\begin{acks}

This work was supported by the UK Research and Innovation Centre for Doctoral Training in Machine Intelligence for Nano-electronic Devices and Systems [EP/S024298/1]. Adam Sobey was supported by The Alan Turing Institute under the Lloyd’s Register Foundation grant ATI/100004. The authors would like to thank Lloyd's Register Foundation for their support, and acknowledge the use of the IRIDIS High-Performance Computing Facility, and associated support services at the University of Southampton, in the completion of this work.

\end{acks}

\printbibliography

\appendix

\section{Experimental Combinations}\label{apx: experimental combinations}

To complement the main experiment description, we include \cref{tab: experimental combinations}, which summarises the different training configurations, evaluated environments, and relevant result sections.

\begin{table}[h]
    \centering
    \caption{Summary of Experimental Combinations. Each architecture is trained with and without the temporal prediction loss, across all values of repetition chance $p \in \{0.25, 0.5, 0.75\}$, in both the Referential Games (RG) and Temporal Referential Games (TRG) training environments. Evaluations are conducted in all six environments. All configurations are run with 10 random seeds. Relevant results are discussed in the sections listed.}
    \label{tab: experimental combinations}
    \adjustbox{max width=\textwidth}{%
        \begin{tabular}{llllll}
            \toprule
            Architecture & Training Env & Temporal Loss & Repetition Chance $p$ & Eval Envs & Results \\
            \midrule
            \basen & RG / TRG & True / False & 0.25 / 0.5 / 0.75 & All 6 & \cref{ssec: temporality analysis} \\
            \temporaln & RG / TRG & True / False & 0.25 / 0.5 / 0.75 & All 6 & \cref{ssec: temporality analysis,sec: discussion} \\
            \temporalrn & RG / TRG & True / False & 0.25 / 0.5 / 0.75 & All 6 & \cref{ssec: temporality analysis,sec: discussion}\\
            \bottomrule
        \end{tabular}
    }
\end{table}

\section{Training Details}\label{apx: training details}

Our agents were trained using PyTorch Lightning \citep{falcon_pytorch_2019} using the Adam optimizer \citep{kingma_adam_2015}, with experiment tracking done via Weights \& Biases \citep{biewald_experiment_2020}. We provide our grid search parameters per network and per training environment in \cref{tab: training params}. We ran a manual grid search over these parameters for each network and training dataset combination, where the networks were \basen, \temporaln, \temporalrn  and the training datasets were Referential Games or Temporal Referential Games. Each trained network was then evaluated on the six available environments: Always Same, Never Same, Referential Games, Temporal Referential Games, Hard Referential Games, and Hard Temporal Referential Games. Running the grid search for one iteration, with the value of repetition chance fixed, took approximately 28 hours, using the compute resources in \cref{tab: compute resources}.

\begin{table}[h]
    \caption{Grid Search Parameters}
    \label{tab: training params}
    \centering
    \begin{tabular}{ll}
        \toprule
        Parameter & Value \\
        \hline
        Epochs & [600] \\
        Optimizer & Adam \\
        Learning Rate $\alpha$ & 0.001 \\
        Number of Objects in Dataset & [20 000] \\
        Number of Distractors & [10] \\
        Number of Attributes $N_{att}$ & [8] \\
        Number of Values $N_{val}$ & [8] \\
        Temporal Prediction Loss Present & [True, False] \\
        Length Penalty & [0] \\
        Maximum Message Length $L$ & [5] \\
        Vocabulary Size $N_{vocab}$ & [26] \\
        Repetition Chance ($p$) & [0.25, 0.5, 0.75] \\
        Previous Horizon $h$ & [8] \\
        Sender Embedding Size & [128] \\
        Sender Meaning LSTM Hidden Size & [128] \\
        Sender Temporal LSTM Hidden Size & [128] \\
        Sender Message LSTM Hidden Size & [128] \\
        Receiver LSTM+Linear Hidden Size & [128] \\
        Gumbel-Softmax Temperature & [1.0] \\
        \bottomrule
    \end{tabular}
\end{table}

\begin{table}[h]
    \caption{Compute Resources}
    \label{tab: compute resources}
    \centering
    \begin{tabular}{ll}
        \toprule
        Resource & Quantity \\
        \hline
        CPU Cores (Intel(R) Xeon(R) Silver 4216 × 2) & 20 \\
        GPUs (NVIDIA Quadro RTX8000) & 1 \\
        Wall Time & 28hrs \\
        \bottomrule
    \end{tabular}
\end{table}

\section{Datasets Details}\label{apx: dataset details}

\begin{figure}[h]
    \begin{center}
    \centerline{\includegraphics[width=0.67\columnwidth]{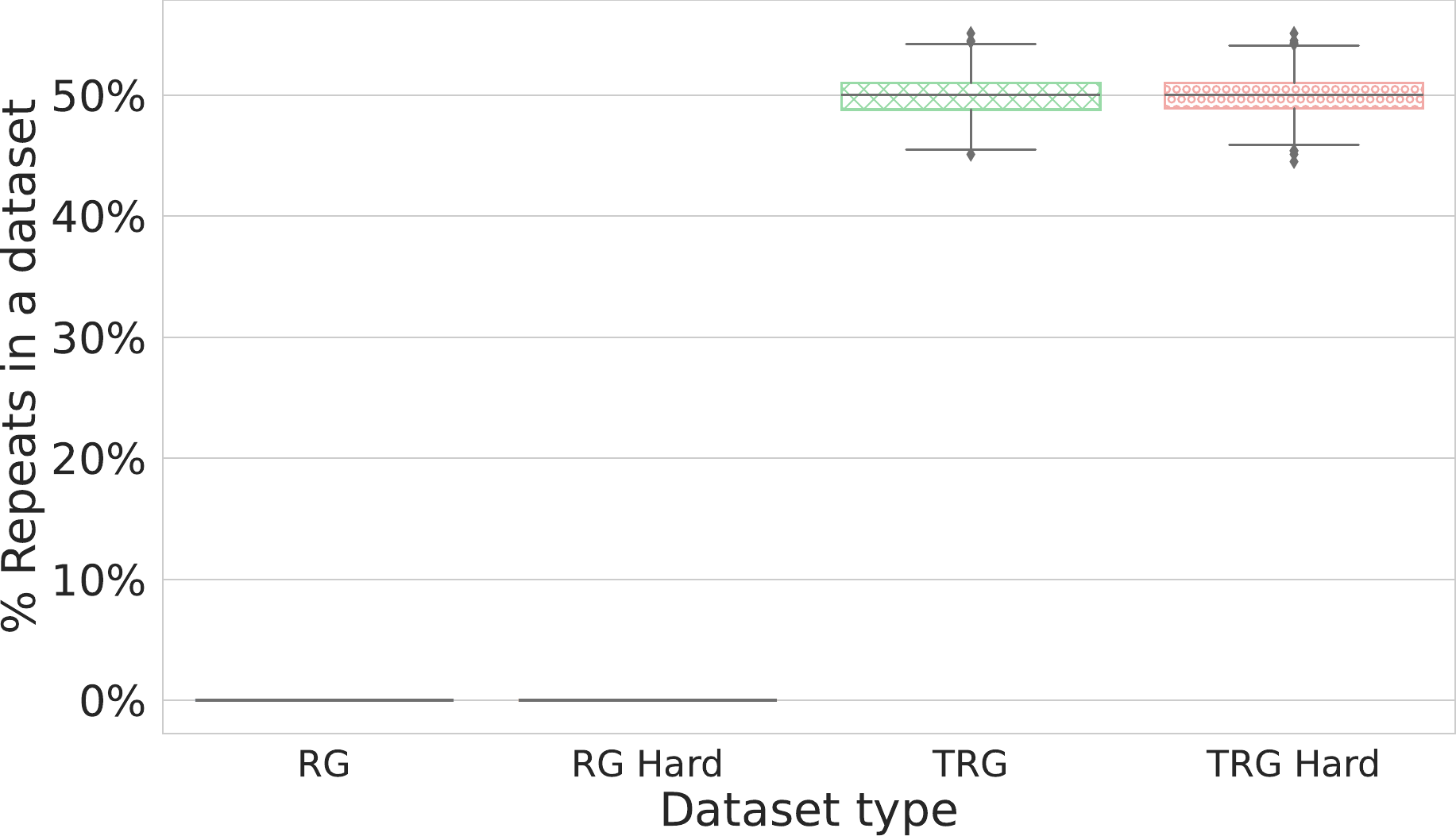}}
    \caption{Number of target repetitions per dataset. Regular referential games datasets very rarely encounter target repetitions. This data is an average over 1000 seeds per environment.}
    \label{fig: dataset repeats}
    \end{center}
\end{figure}

In \cref{fig: dataset repeats}, we analyse our datasets, using the parameters as specified in \cref{apx: training details}, for the number of repetitions that occur. When the temporal dataset repetition chance is set to 50\%, the datasets, predictably, oscillate around 50\% of repeating targets. Generating the targets randomly yields a miniscule fraction of repetitions of less than 1\%, as we can see in \cref{fig: dataset repeats}, for the Classic and Hard referential games.

\subsection{Test Environments}\label{apx: test environments}

Both Always Same and Never Same environments act as sanity checks for our results.

We provide example inputs and outputs for both environments in \cref{tab: example messages as} and \cref{tab: example messages ns}. We use single-attribute objects and messages for clarity.

For the Always Same environment, in the case of the agent using temporal references, we may also see other messages instead of the message $\bm{m}_4$, as we have observed that there are more than one message used as previously. We always expect to see at most 90\% of usage as previously for this environment, unless the agents learn temporal referencing strategies, when we would expect the usage to reach 100\%.

For the Never Same environment, we expect to see no temporal references being identified. Any identification of temporal references in the Never Same environment would indicate an issue with our metric.

\begin{table}[h]
    \centering
    \caption{Example Inputs and Outputs for Always Same.}
    \begin{tabular}{c|cc}
         Environment & Always Same \\
         \hline
         Input & $[\bm{x},\bm{x},\bm{x},\bm{y},\bm{y},\bm{y},\bm{z},\bm{z},\bm{z}]$ \\
         Temporal Referencing & $[\bm{m}_1,\bm{m}_4,\bm{m}_4,\bm{m}_2,\bm{m}_4,\bm{m}_4,\bm{m}_3,\bm{m}_4,\bm{m}_4]$ \\
         No Temporal Referencing & $[\bm{m}_1,\bm{m}_2,\bm{m}_3,\bm{m}_4,\bm{m}_5,\bm{m}_6,\bm{m}_7,\bm{m}_8]$ \\
    \end{tabular}
    \label{tab: example messages as}
\end{table}

\begin{table}[h]
    \centering
    \caption{Example Inputs and Outputs for Never Same.}
    \begin{tabular}{c|cc}
         Environment & Never Same \\
         \hline
         Input & $[\bm{x},\bm{y},\bm{z},\bm{a},\bm{b},\bm{c},\bm{d},\bm{e}]$ \\
         Temporal Referencing & $[\bm{m}_1,\bm{m}_1,\bm{m}_1,\bm{m}_2,\bm{m}_2,\bm{m}_2,\bm{m}_3,\bm{m}_3,\bm{m}_3]$ \\
         No Temporal Referencing & $[\bm{m}_1,\bm{m}_2,\bm{m}_3,\bm{m}_4,\bm{m}_5,\bm{m}_6,\bm{m}_7,\bm{m}_8]$ \\
    \end{tabular}
    \label{tab: example messages ns}
\end{table}

\section{Accuracy Distributions}\label{apx: accuracy}

\begin{figure}[!ht]
    \centering
    \begin{subfigure}{0.475\textwidth}
        \centering
        \includegraphics[width=\textwidth]{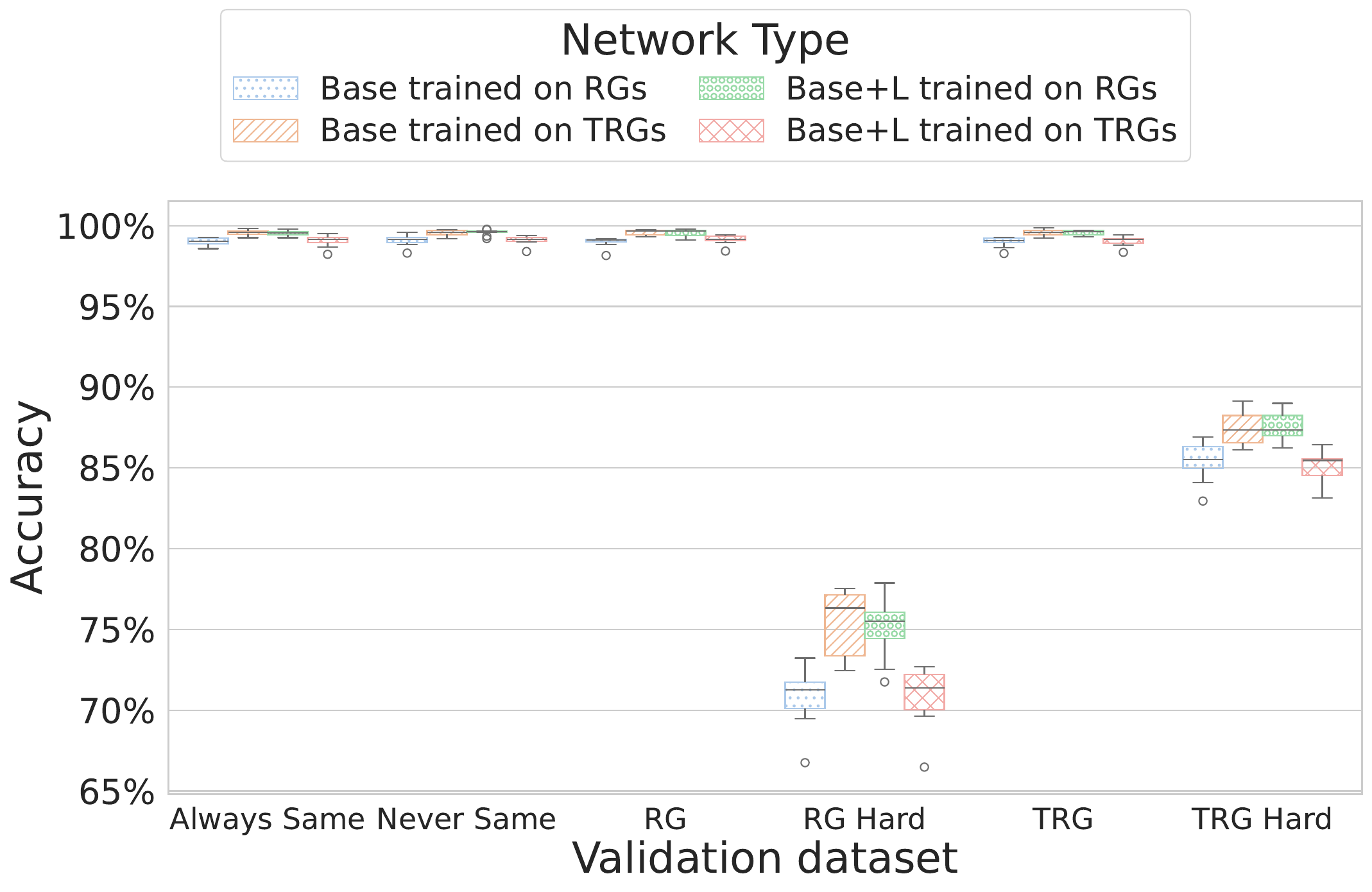}
        \caption{The evaluation accuracy for the \basen agents across all environments.}
        \label{fig: base acc}
    \end{subfigure}
    \hfill
    \begin{subfigure}{0.475\textwidth}
        \centering
        \includegraphics[width=\textwidth]{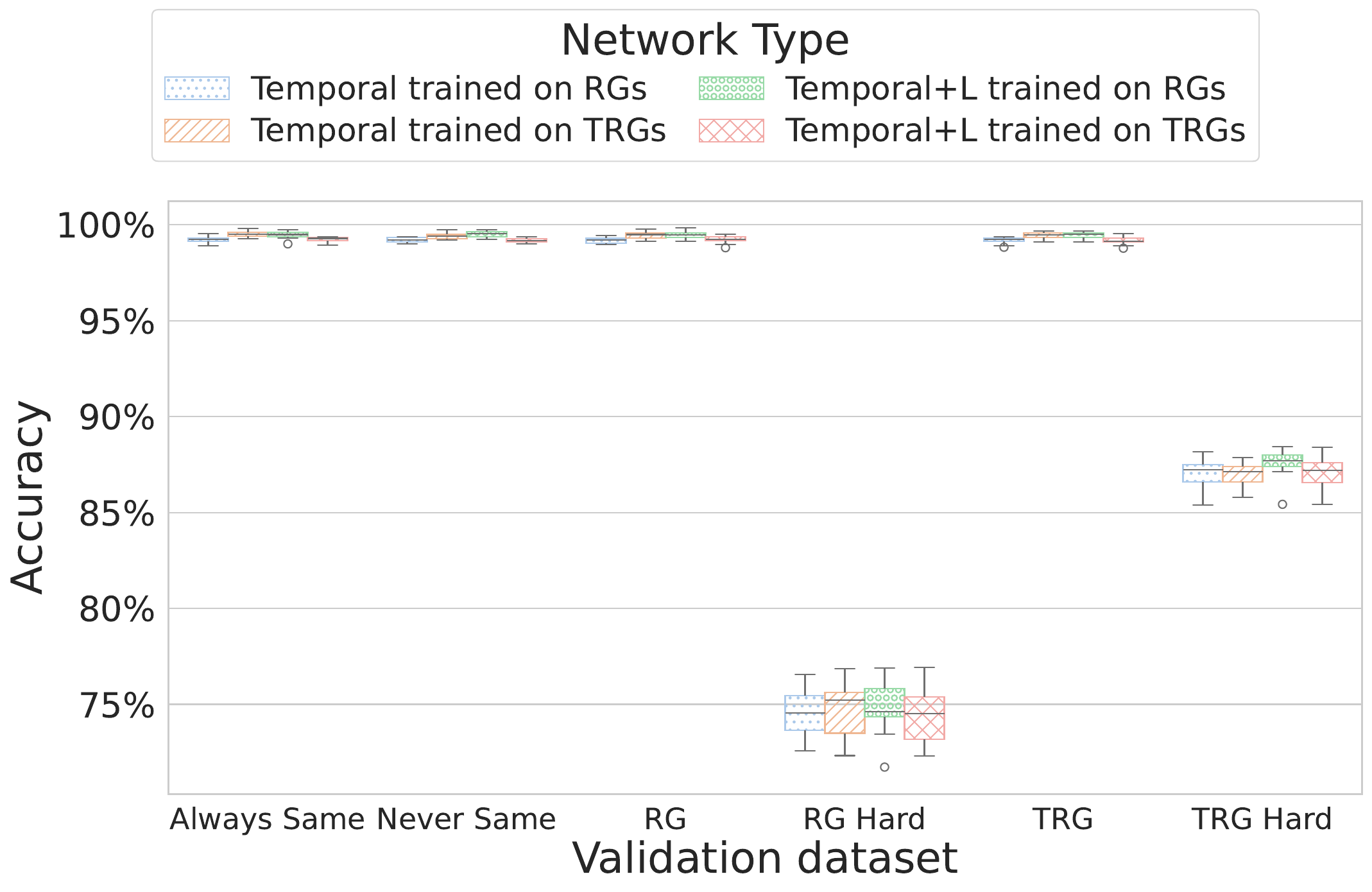}
        \caption{The evaluation accuracy for the \temporaln  agents across all environments.}
        \label{fig: temporal acc}
    \end{subfigure}
    \vskip\baselineskip
    \begin{subfigure}{0.475\textwidth}
        \centering
        \includegraphics[width=\textwidth]{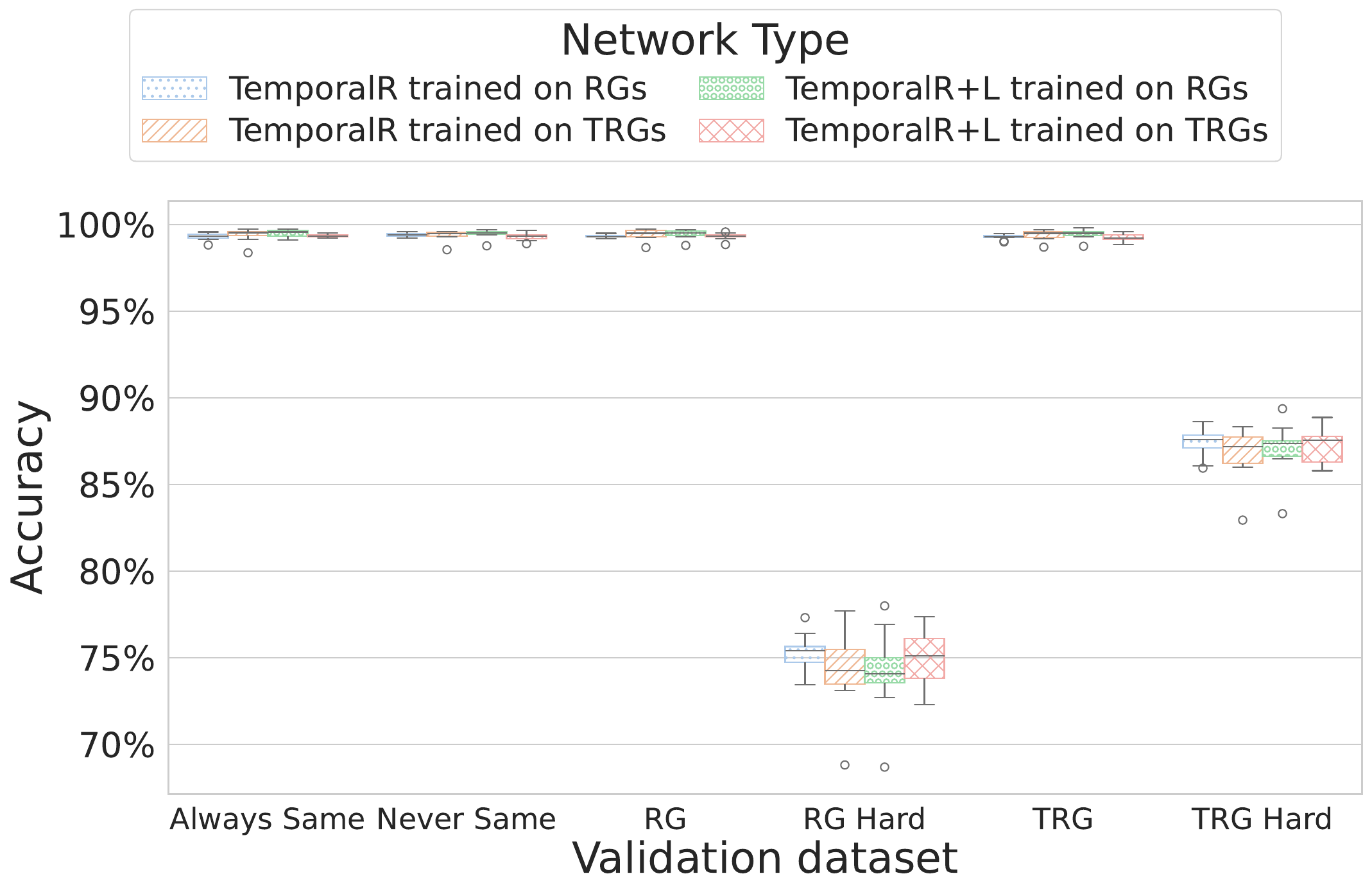}
        \caption{The evaluation accuracy for the \temporalrn  agents across all environments.}
        \label{fig: temporalr acc}
    \end{subfigure}
    \hfill
    \caption{Accuracies for each network variant on all evaluation environments.}
\end{figure}

The accuracy distributions for all agent types across all evaluation environments are shown in \cref{fig: base acc,fig: temporal acc,fig: temporalr acc}. All agents converge to very similar levels of accuracy.

\clearpage

\section{Topographic Similarity Distributions}\label{apx: topsim}

\begin{figure}[h!]
    \centering
    \begin{subfigure}{0.475\textwidth}
        \centering
        \includegraphics[width=\textwidth]{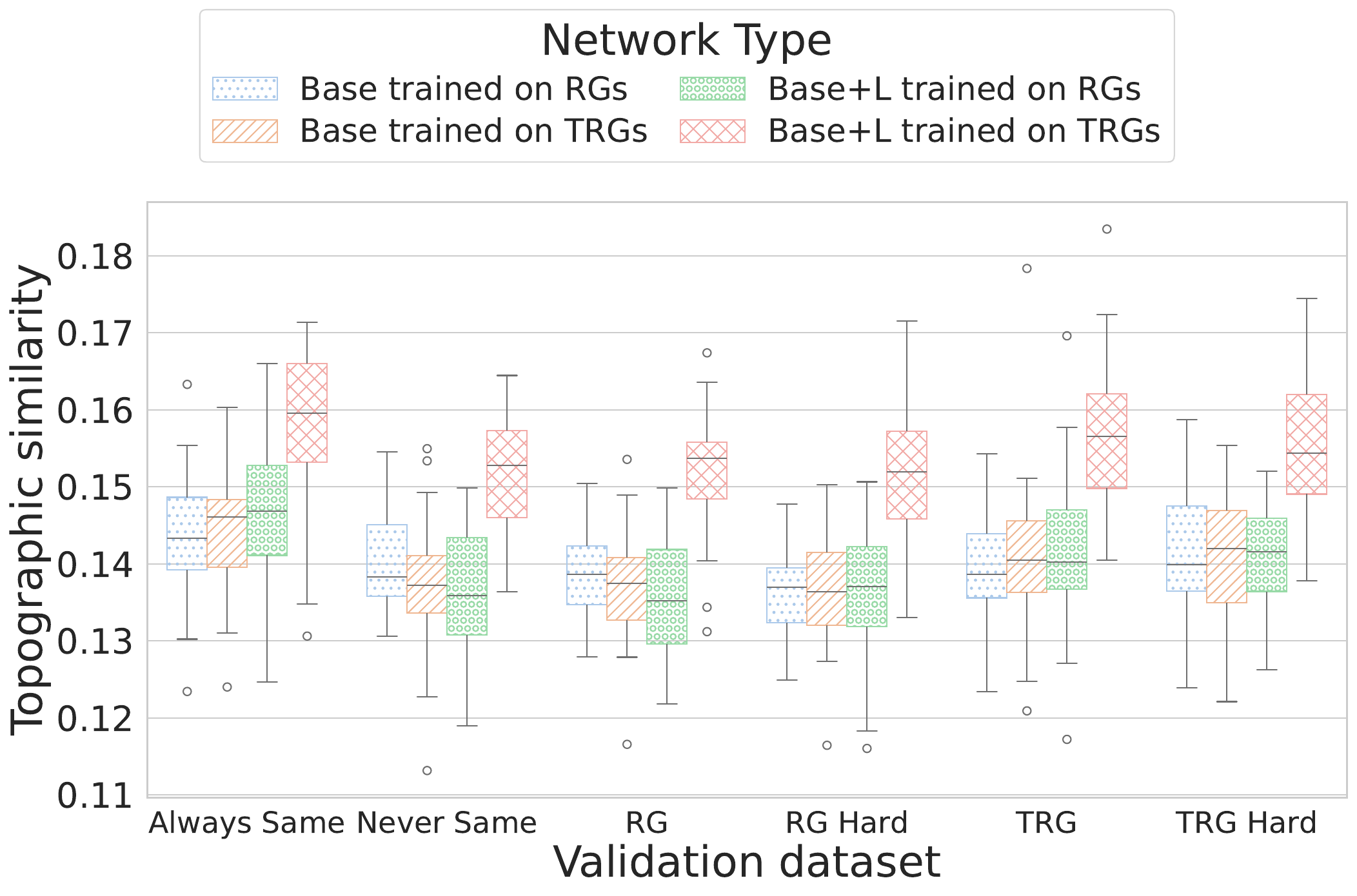}
        \caption{The topographic similarity scores for the \basen agents across all environments.}
        \label{fig: base topsim}
    \end{subfigure}
    \hfill
    \begin{subfigure}{0.475\textwidth}
        \centering
        \includegraphics[width=\textwidth]{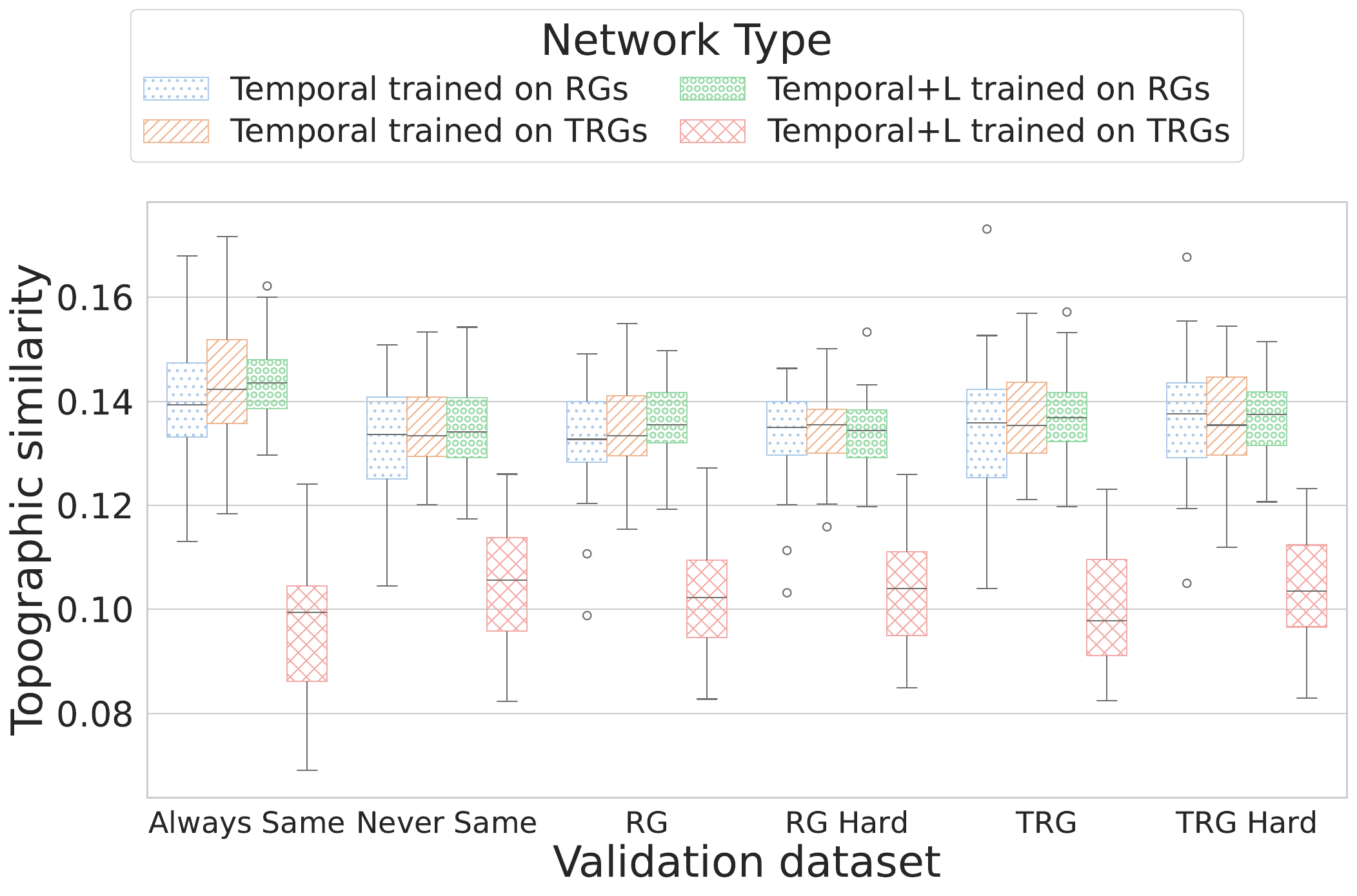}
        \caption{The topographic similarity scores for the \temporaln  agents across all environments.}
        \label{fig: temporal topsim}
    \end{subfigure}
    \vskip\baselineskip
    \begin{subfigure}{0.475\textwidth}
        \centering
        \includegraphics[width=\textwidth]{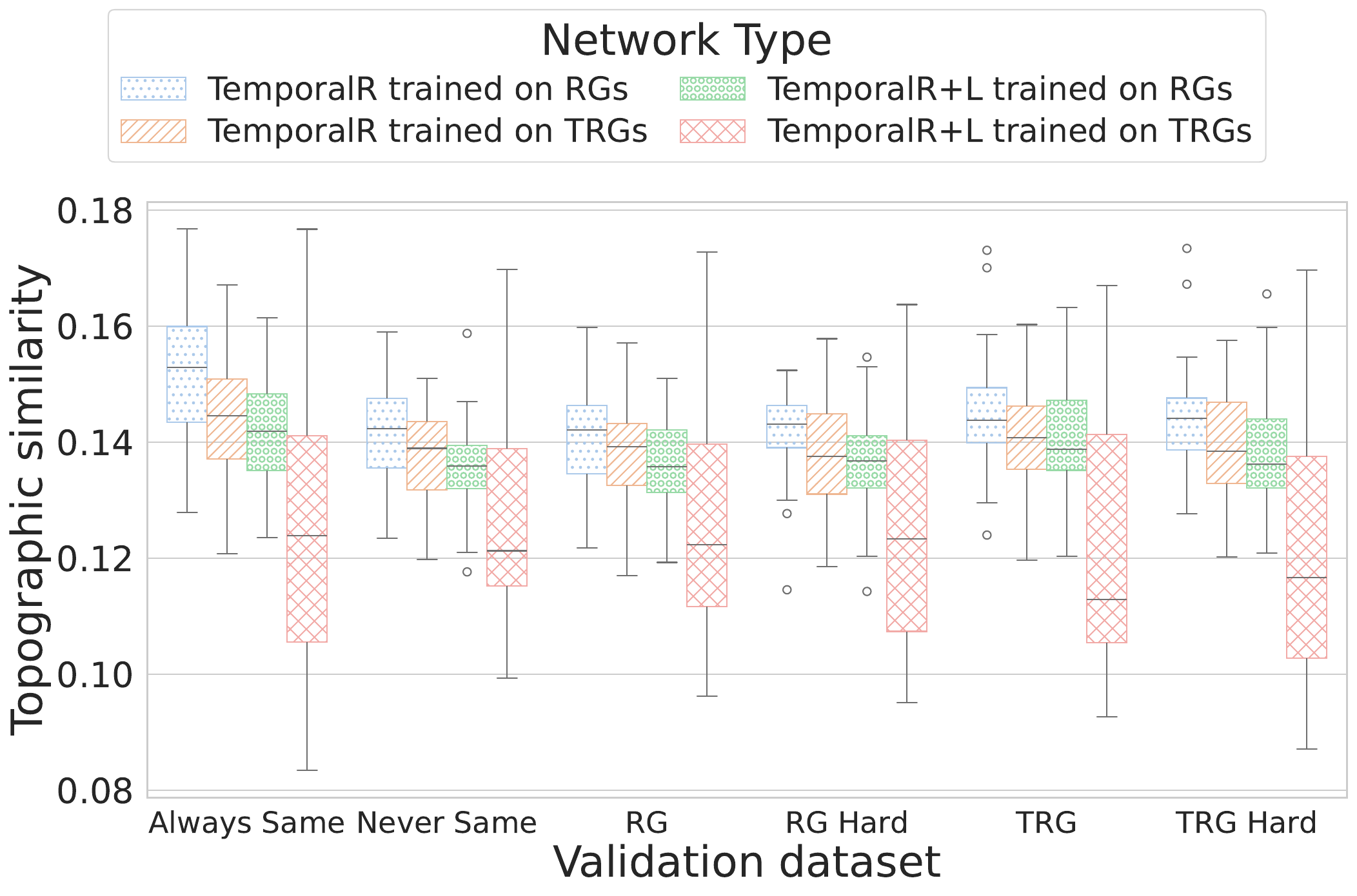}
        \caption{The topographic similarity scores for the \temporalrn  agents across all environments.}
        \label{fig: temporalr topsim}
    \end{subfigure}
    \hfill
    \caption{Topographic similarity scores for each network variant on all evaluation environments.}
\end{figure}

The topographic similarity distributions for all agent types across all evaluation environments are shown in \cref{fig: base topsim,fig: temporal topsim,fig: temporalr topsim}. All agents converge to very similar values of topographic similarity.

\clearpage

\section{Architectural Details}\label{apx: arch details}

\begin{figure}[h]
    \centering
    \begin{subfigure}{0.48\textwidth}
        \centering
        \includegraphics[width=0.9\textwidth]{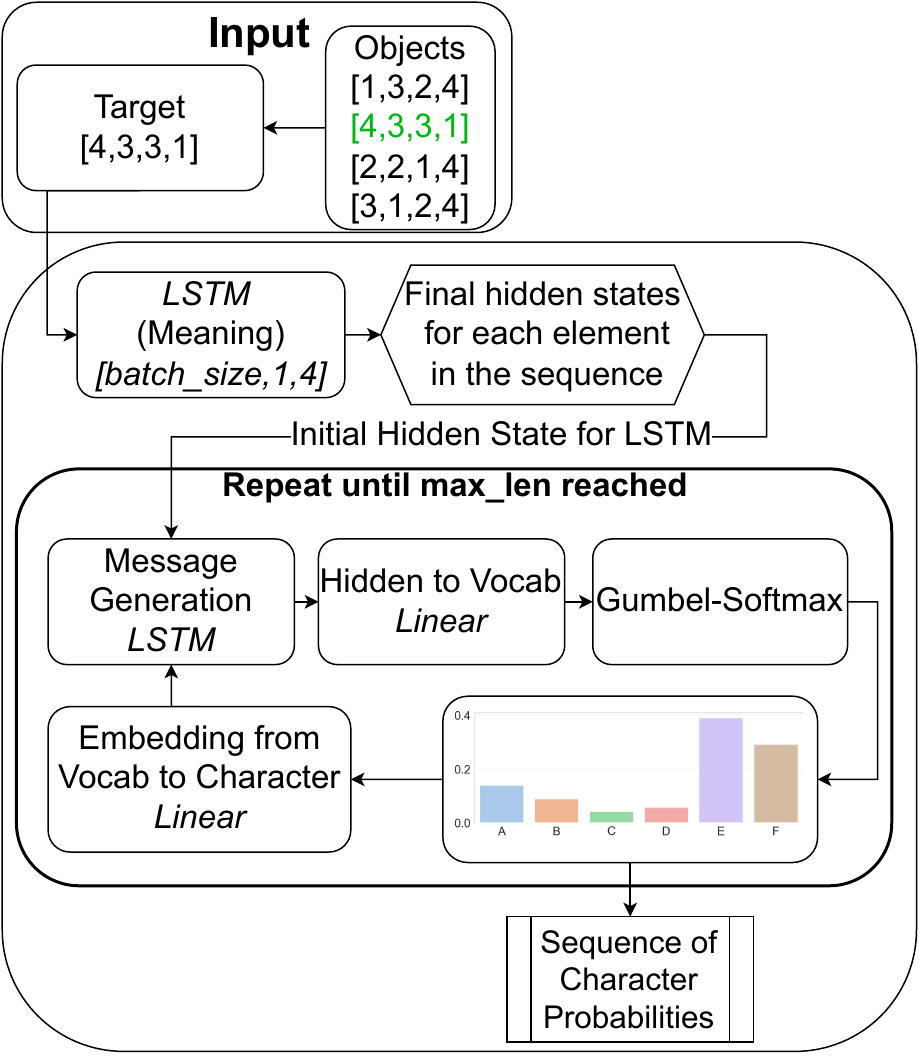}
        \caption{The \basen LSTM sender architecture.}
        \label{fig: arch-sender-base-lstm}
    \end{subfigure}
    \hfill
    \begin{subfigure}{0.48\textwidth}
        \centering
        \includegraphics[width=0.83\textwidth]{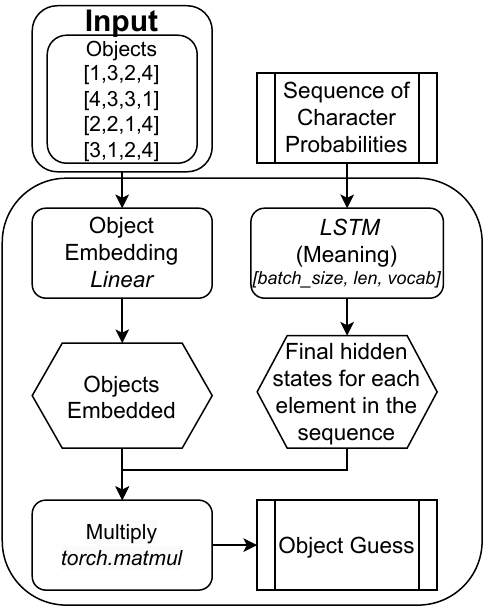}
        \caption{The \basen LSTM receiver architecture.}
        \label{fig: arch-receiver-base-lstm}
    \end{subfigure}
    \caption{The \basen LSTM sender and receiver architectures.}
    \label{fig: arch-base-lstm}
\end{figure}

\begin{figure}[h]
    \centering
    \begin{subfigure}{0.48\textwidth}
        \centering
        \includegraphics[width=0.9\textwidth]{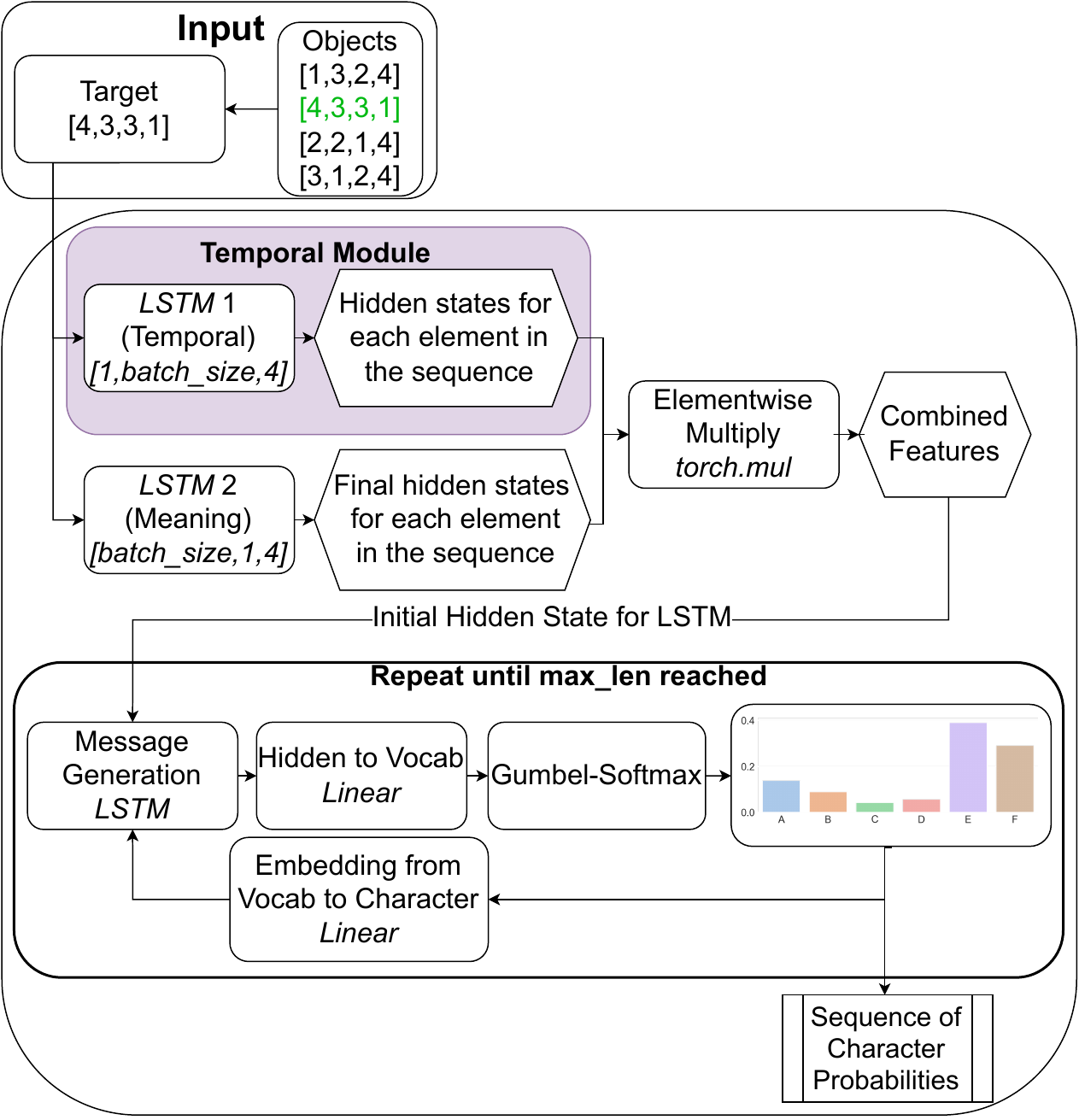}
        \caption{The \temporalrn  LSTM sender architecture.}
        \label{fig: arch-sender-tr-lstm}
    \end{subfigure}
    \hfill
    \begin{subfigure}{0.48\textwidth}
        \centering
        \includegraphics[width=0.9\textwidth]{figures/TRGL_LSTM_Receiver.drawio.pdf}
        \caption{The \temporalrn  LSTM receiver architecture.}
        \label{fig: arch-receiver-tr-lstm}
    \end{subfigure}
    \caption{The \temporalrn  LSTM sender and receiver architectures, with the temporal modules highlighted in purple.}
    \label{fig: arch-tr-lstm}
\end{figure}

In \cref{fig: arch-base-lstm} and \cref{fig: arch-tr-lstm} we present the detailed overview of the \basen and \temporalrn agents respectively. Compared to the \temporaln agent, the \basen agent lacks the sequential LSTM, while the \temporalrn agent has the regularly batched LSTM, in addition to the sequential LSTM.

\begin{figure}[h]
    \begin{center}
    \centerline{\includegraphics[width=0.9\columnwidth]{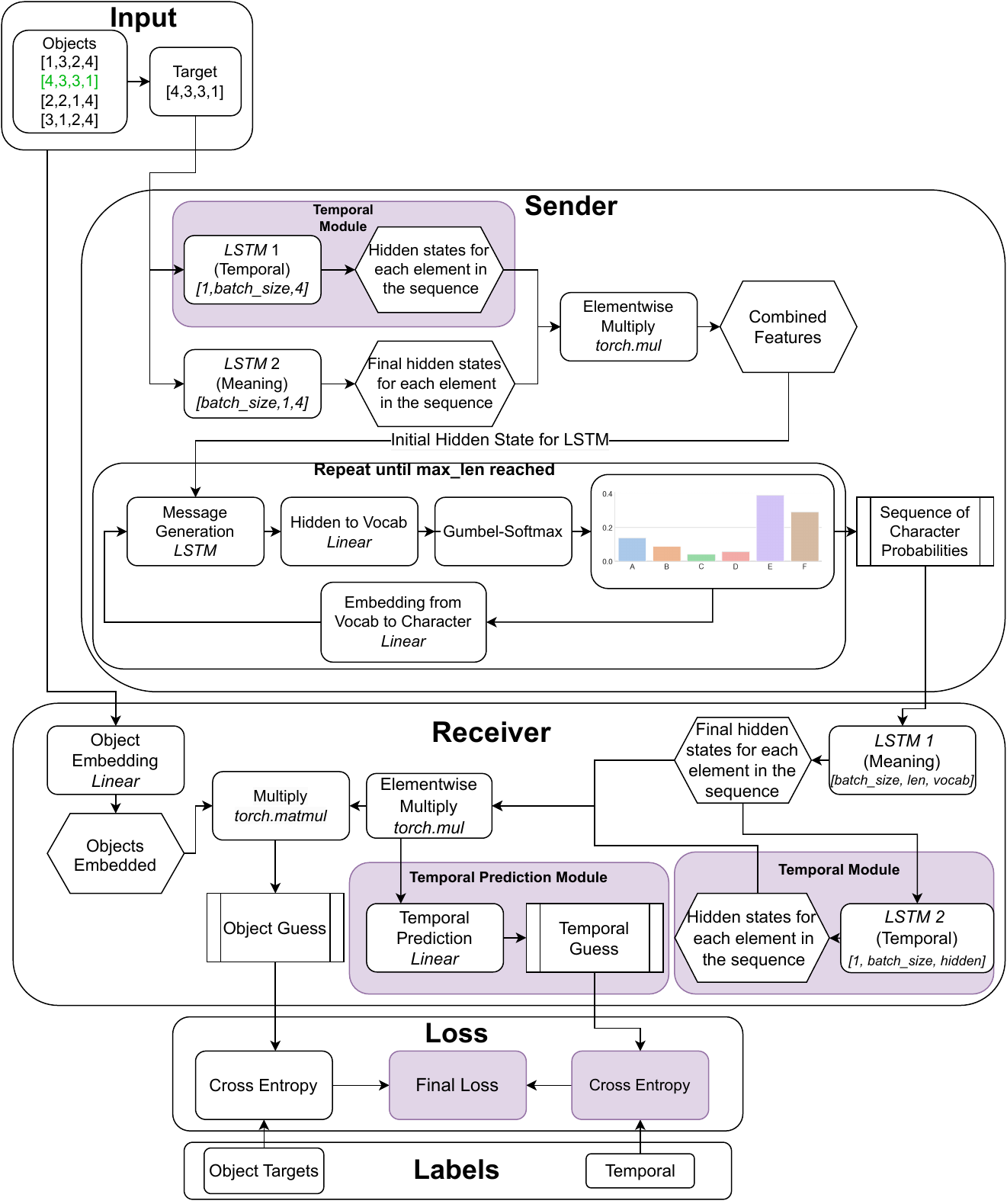}}
    \caption{Full overview of our Temporal Referential Games setup. Together with the sender and receiver we have described in \cref{sec: architectures}, we also include the working of the loss.}
    \label{fig: trg full overview lstm}
    \end{center}
\end{figure}

In \cref{fig: trg full overview lstm}, we present an overview of our whole experimental setup. We can see the sender and receiver architectures, together with their inputs, as described in \cref{sec: architectures}. We also show our loss calculations. 

\clearpage

\clearpage

\section{Supplementary Material}

\subsection{Temporality analysis for previous horizon from \texorpdfstring{$h_v=1$}{hv=1} to \texorpdfstring{$h_v=8$}{hv=8}}\label{apx: horizon 8}

In \cref{tab: perc emergence h8}, we can see the absolute percentages of networks which develop temporal references.

\begin{figure}[t]
	\centering
	\includegraphics[width=0.9\textwidth]{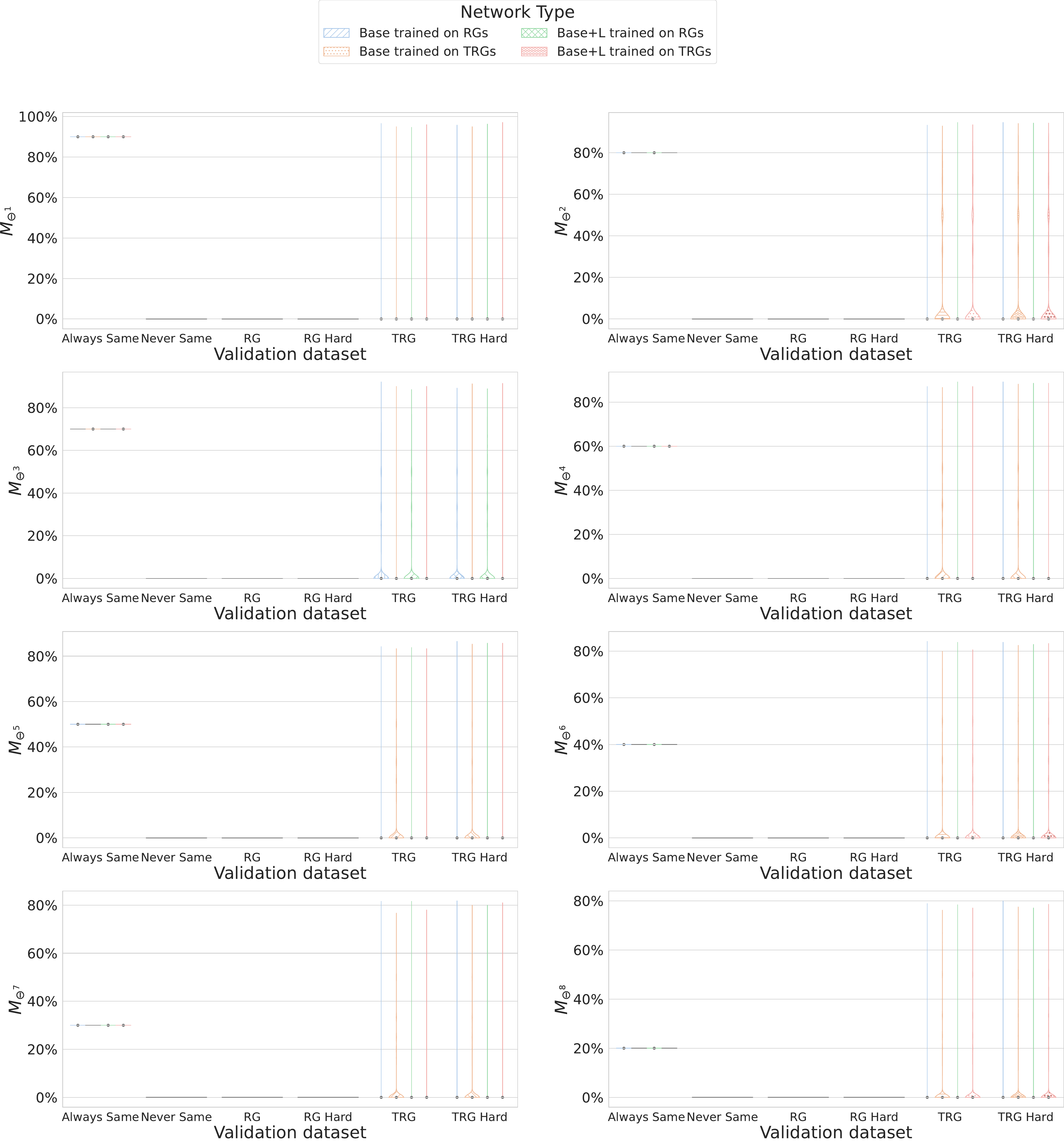}
	\caption{The $M_{\ominus^h}$ metric values per message for the \textit{Base} agents, for all environments.}
	\label{fig: prev usage h8 base}
\end{figure}

\begin{figure}[t]
	\centering
	\includegraphics[width=0.9\textwidth]{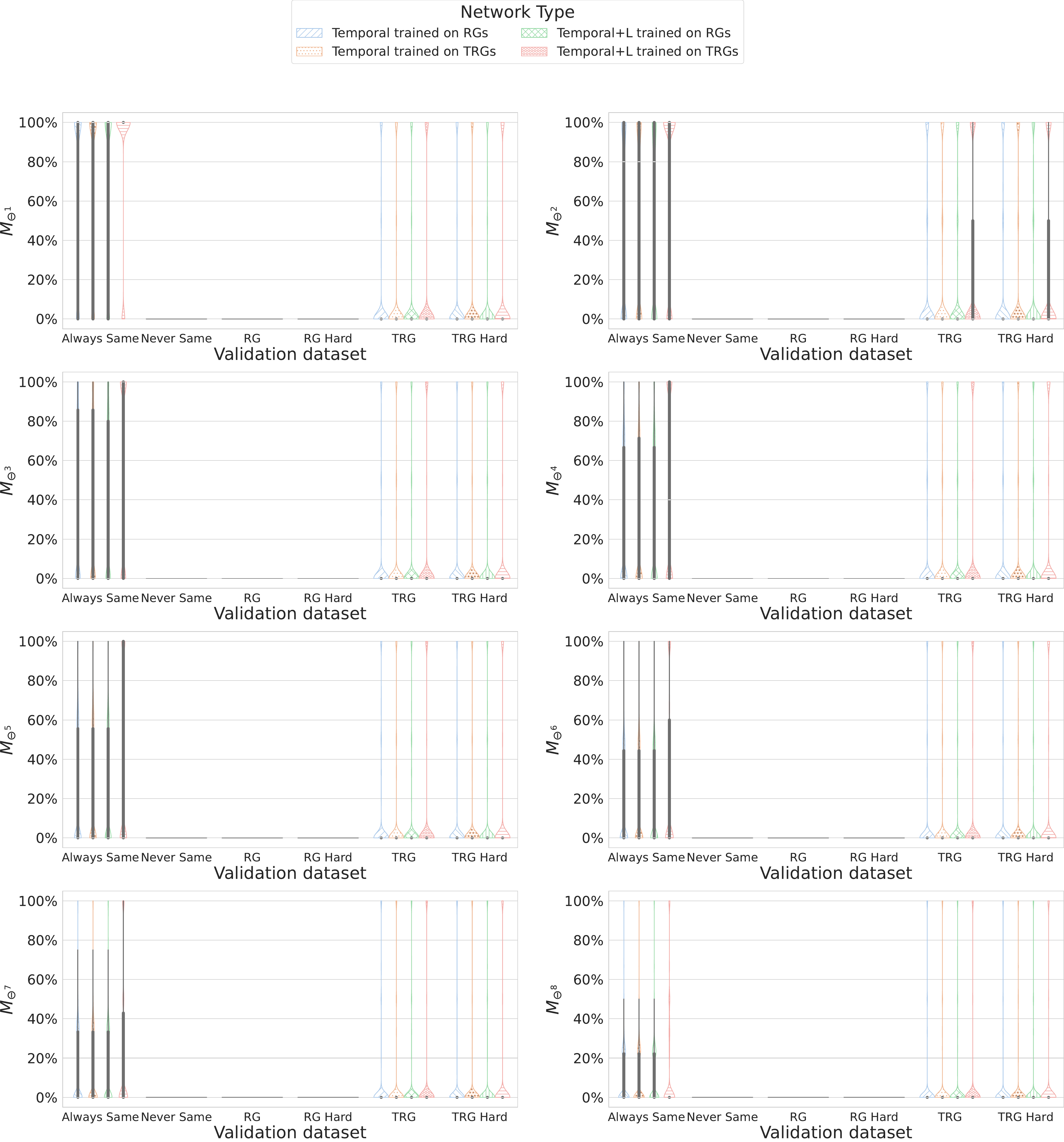}
	\caption{The $M_{\ominus^h}$ metric values per message for the \textit{Temporal} agents, for all environments.}
	\label{fig: prev usage h8 temporal}
\end{figure}

\begin{figure}[t]
	\centering
	\includegraphics[width=0.9\textwidth]{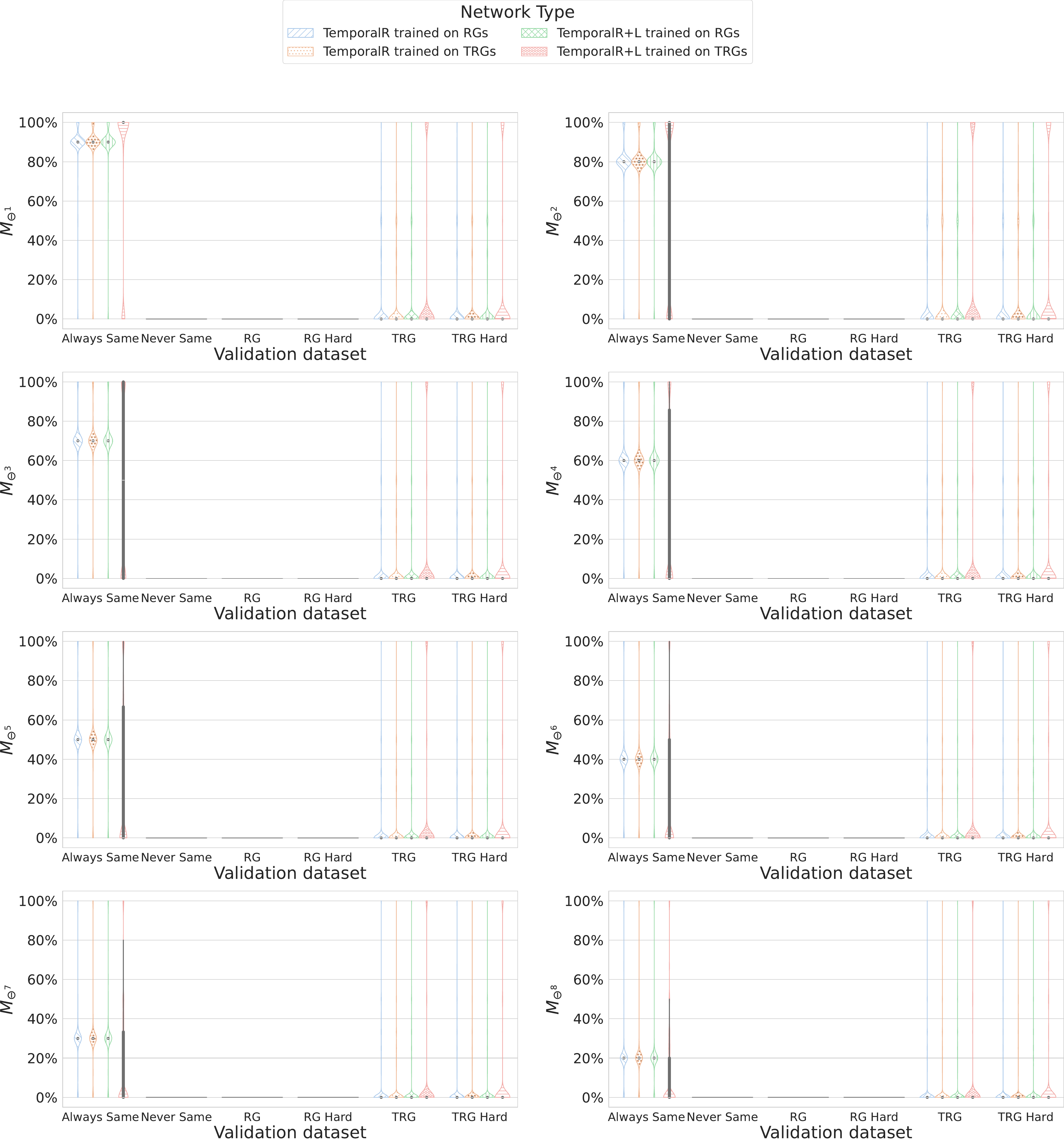}
	\caption{The $M_{\ominus^h}$ metric values per message for the \textit{TemporalR} agents, for all environments.}
	\label{fig: prev usage h8 temporalr}
\end{figure}

\begin{figure}[h]
	\centering
	\includegraphics[width=0.9\textwidth]{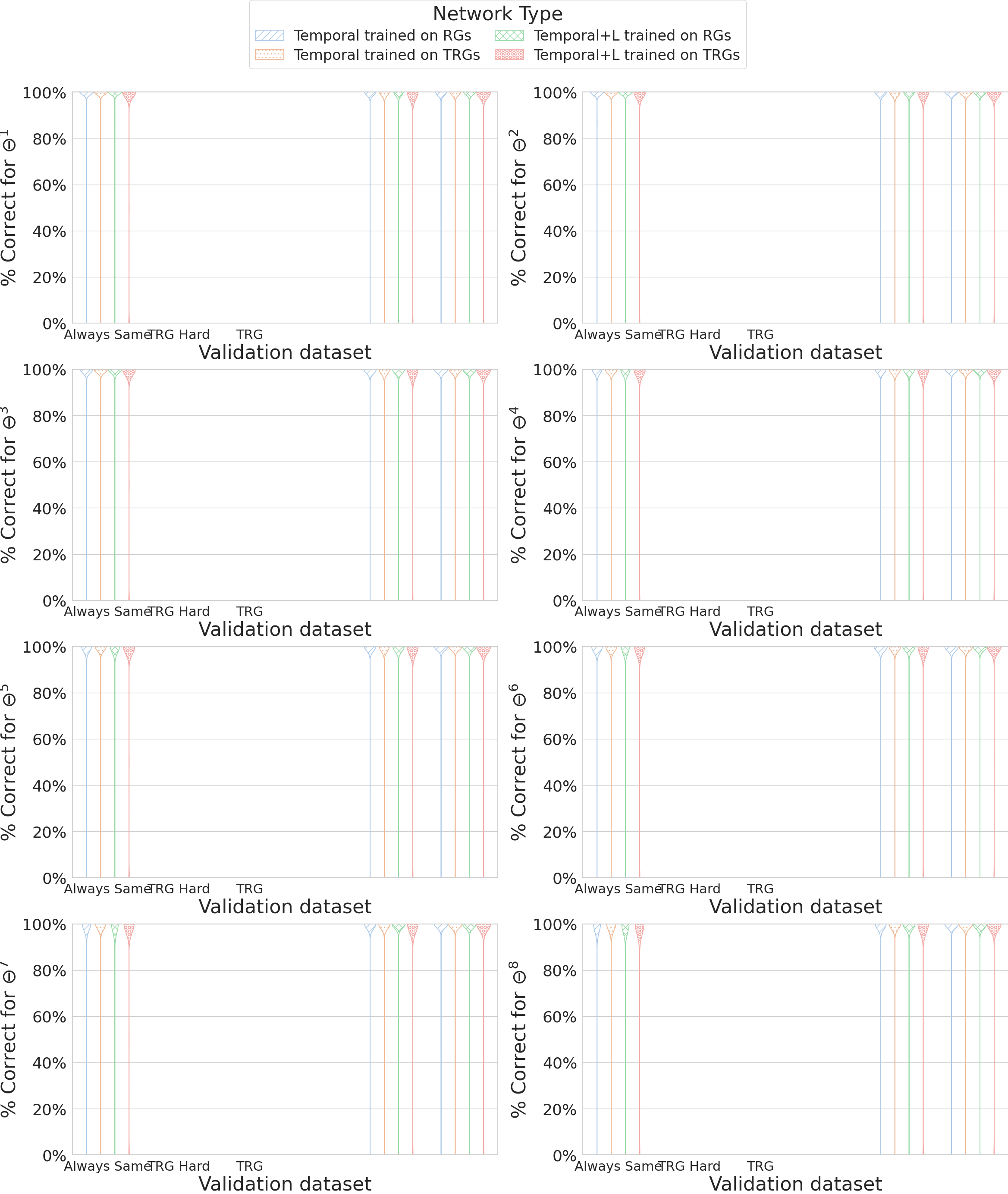}
	\caption{Correctness of messages used as the $\ominus^h$ operator for the \textit{Temporal} agents.}
	\label{fig: prev correct h8 temporal}
\end{figure}

\begin{figure}[h]
	\centering
	\includegraphics[width=0.9\textwidth]{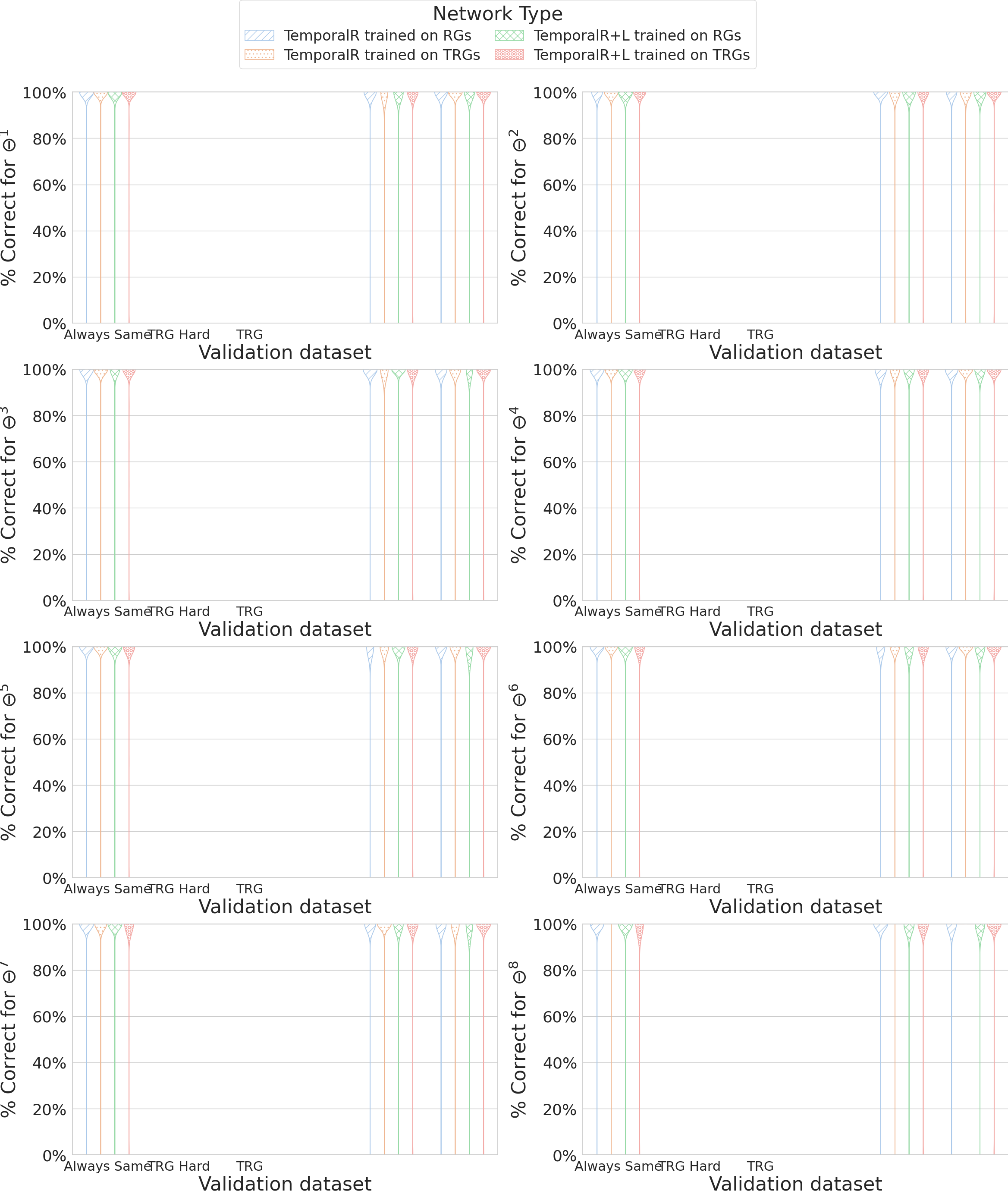}
	\caption{Correctness of messages used as the $\ominus^h$ operator for the \textit{TemporalR} agents.}
	\label{fig: prev correct h8 temporalr}
\end{figure}

The full plots for all metrics are shown for previous horizon from \(h_v=1\) to \(h_v=8\). The $M_{\ominus^h}$ metric values for the Base, Temporal, and TemporalR agents are available in \cref{fig: prev usage h8 base,fig: prev usage h8 temporal,fig: prev usage h8 temporalr}. The correctness of the messages used as the $\ominus^h$ operator for each agent type is available in \cref{fig: prev correct h8 temporal,fig: prev correct h8 temporalr}. No Base networks feature in the plots, as they do not develop temporal references, and so there are no messages to measure the correctness of.

\begin{table}[h]
	\centering
	\small
	\caption{Emergence of temporal references for a given horizon.}
	\begin{tabular}{lllllllll}
		Network Type & $h_v=1$ & $h_v=2$ & $h_v=3$ & $h_v=4$ & $h_v=5$ & $h_v=6$ & $h_v=7$ & $h_v=8$ \\
		\toprule
		Base         & 0\%     & 0\%     & 0\%     & 0\%     & 0\%     & 0\%     & 0\%     & 0\%     \\
		Base+L       & 0\%     & 0\%     & 0\%     & 0\%     & 0\%     & 0\%     & 0\%     & 0\%     \\
		\midrule
		Temporal     & 100\%   & 100\%   & 100\%   & 100\%   & 100\%   & 100\%   & 100\%   & 100\%   \\
		Temporal+L   & 100\%   & 100\%   & 100\%   & 100\%   & 100\%   & 100\%   & 100\%   & 100\%   \\
		\midrule
		TemporalR    & 98.89\% & 100\%   & 100\%   & 99.44\% & 97.22\% & 100\%   & 99.44\% & 97.78\% \\
		TemporalR+L  & 99.44\% & 100\%   & 99.44\% & 98.89\% & 98.89\% & 99.44\% & 99.44\% & 98.89\% \\
	\end{tabular}
	\label{tab: perc emergence h8}
\end{table}

\begin{figure}[h]
	\centering
	\includegraphics[width=0.85\textwidth]{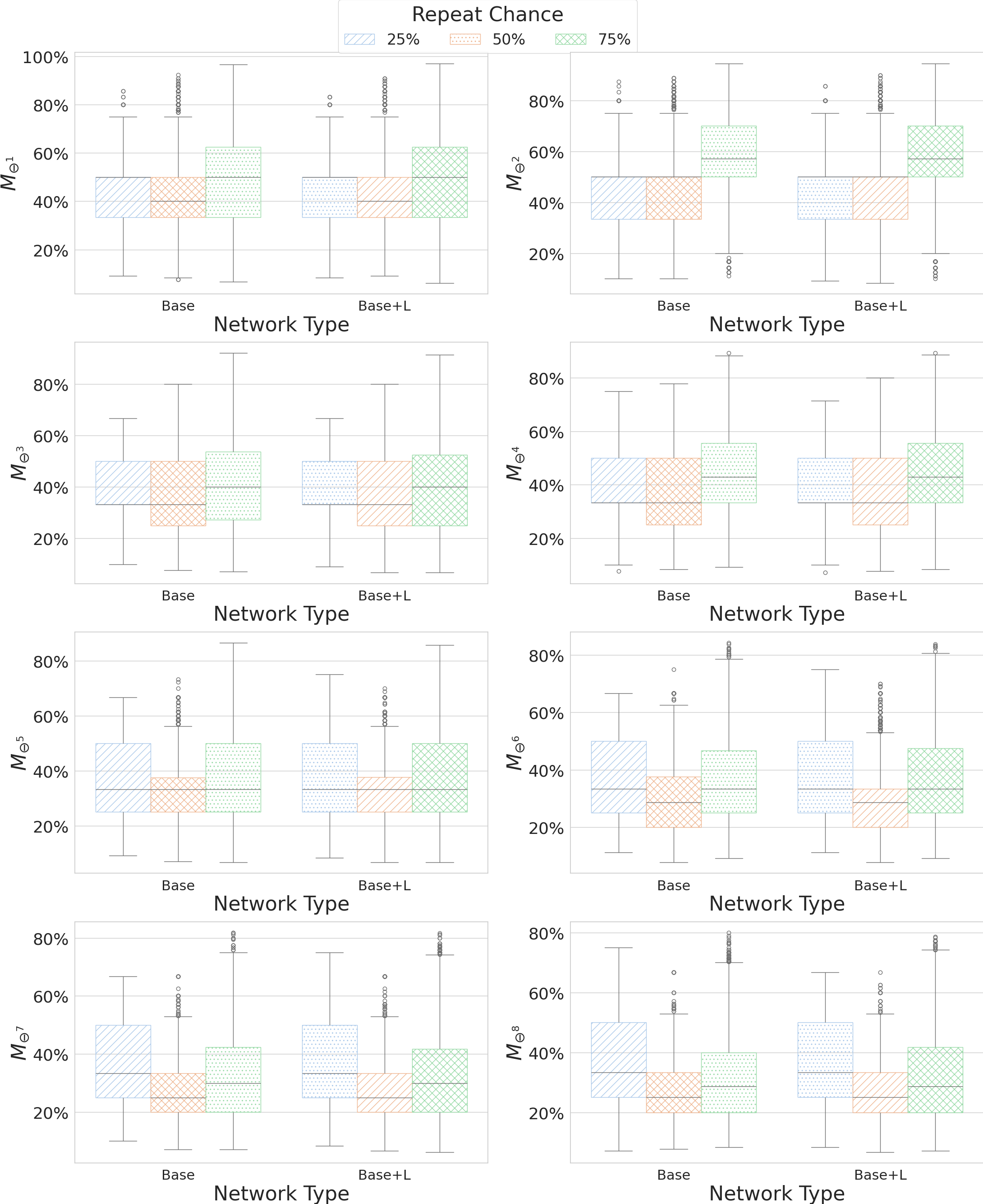}
	\caption{The $M_{\ominus^h}$ value when varying the loss and the chance of repetition for the \textit{Base} agents.}
	\label{fig: repeat chance h8 base}
\end{figure}

\begin{figure}[h]
	\centering
	\includegraphics[width=0.85\textwidth]{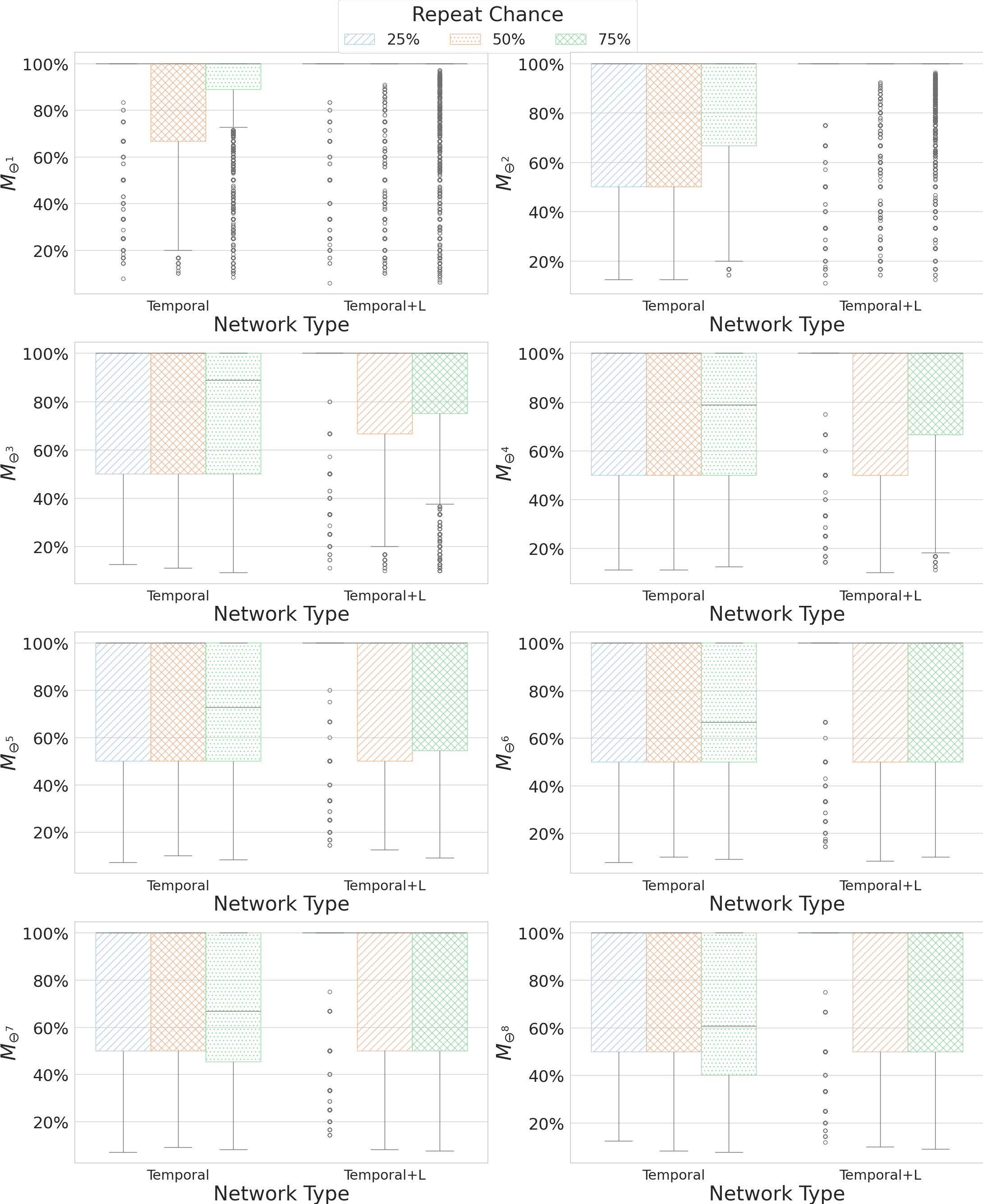}
	\caption{The $M_{\ominus^h}$ value when varying the loss and the chance of repetition for the \textit{Temporal} agents.}
	\label{fig: repeat chance h8 temporal}
\end{figure}

\begin{figure}[h]
	\centering
	\includegraphics[width=0.85\textwidth]{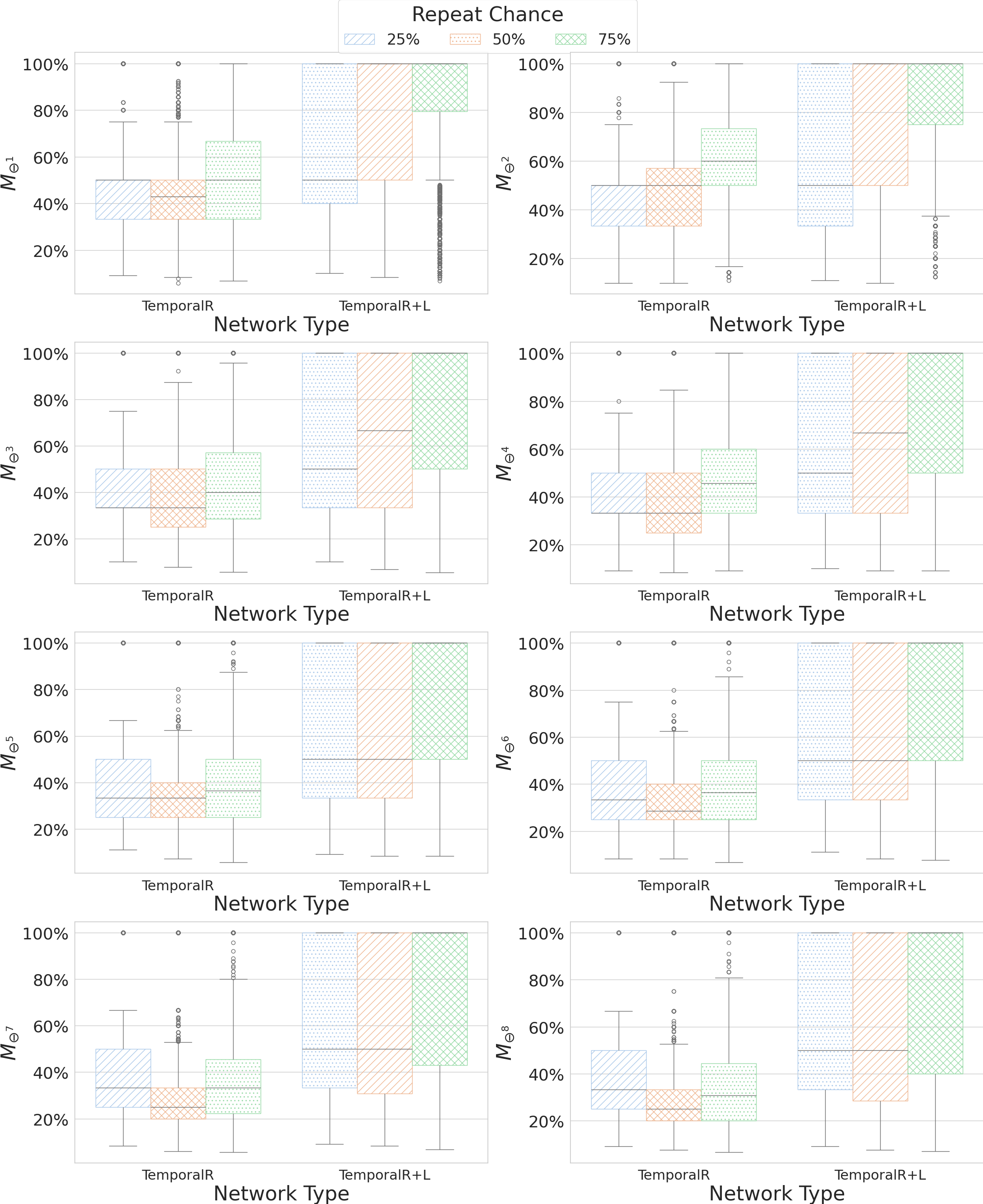}
	\caption{The $M_{\ominus^h}$ value when varying the loss and the chance of repetition for the \textit{TemporalR} agents.}
	\label{fig: repeat chance h8 temporalr}
\end{figure}

\begin{figure}[h]
	\centering
	\includegraphics[width=0.70\textwidth]{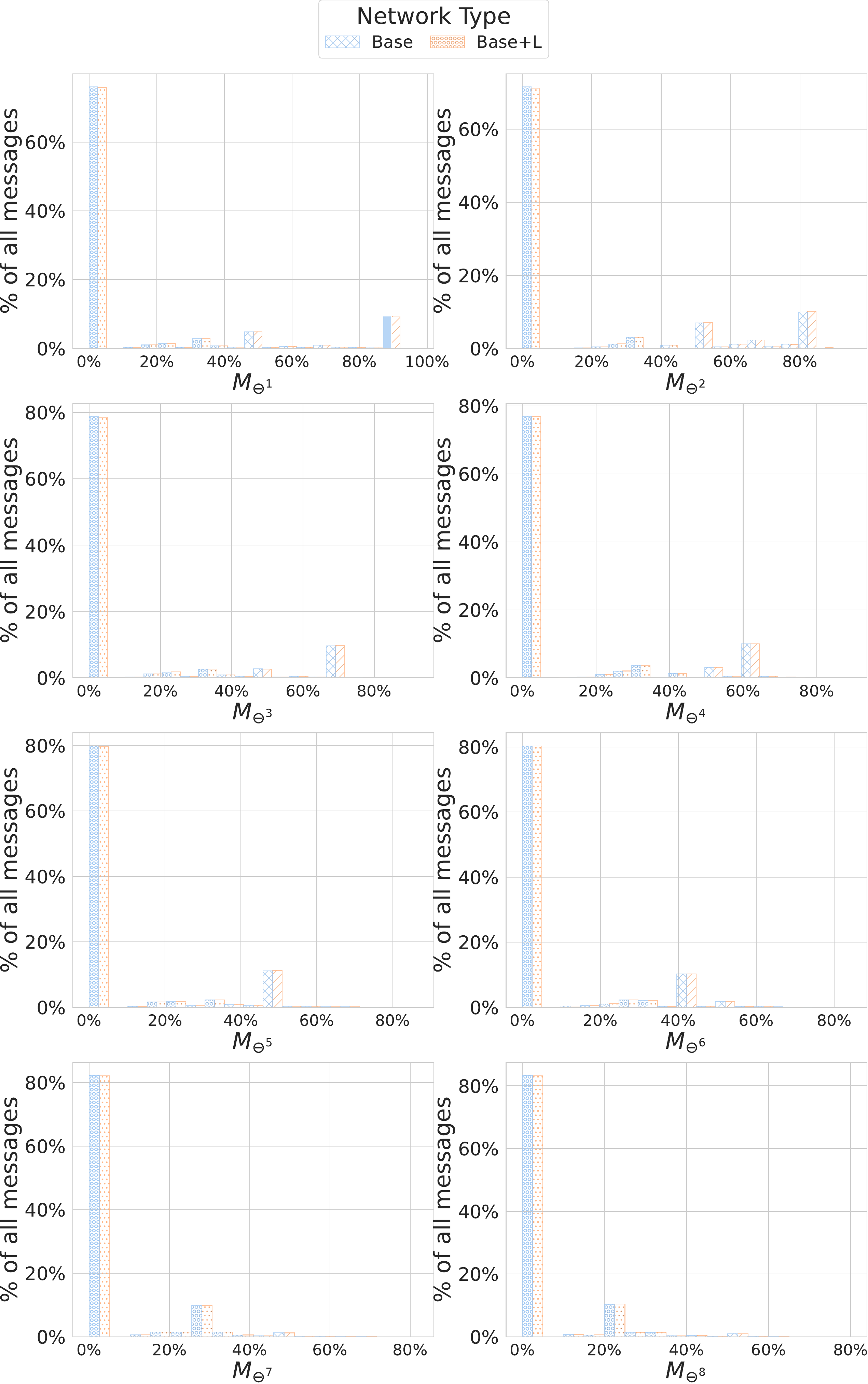}
	\caption{Usage of messages compared to their $M_{\ominus^h}$ value for the \textit{Base} agents.}
	\label{fig: histogram h8 base}
\end{figure}

\begin{figure}[h]
	\centering
	\includegraphics[width=0.70\textwidth]{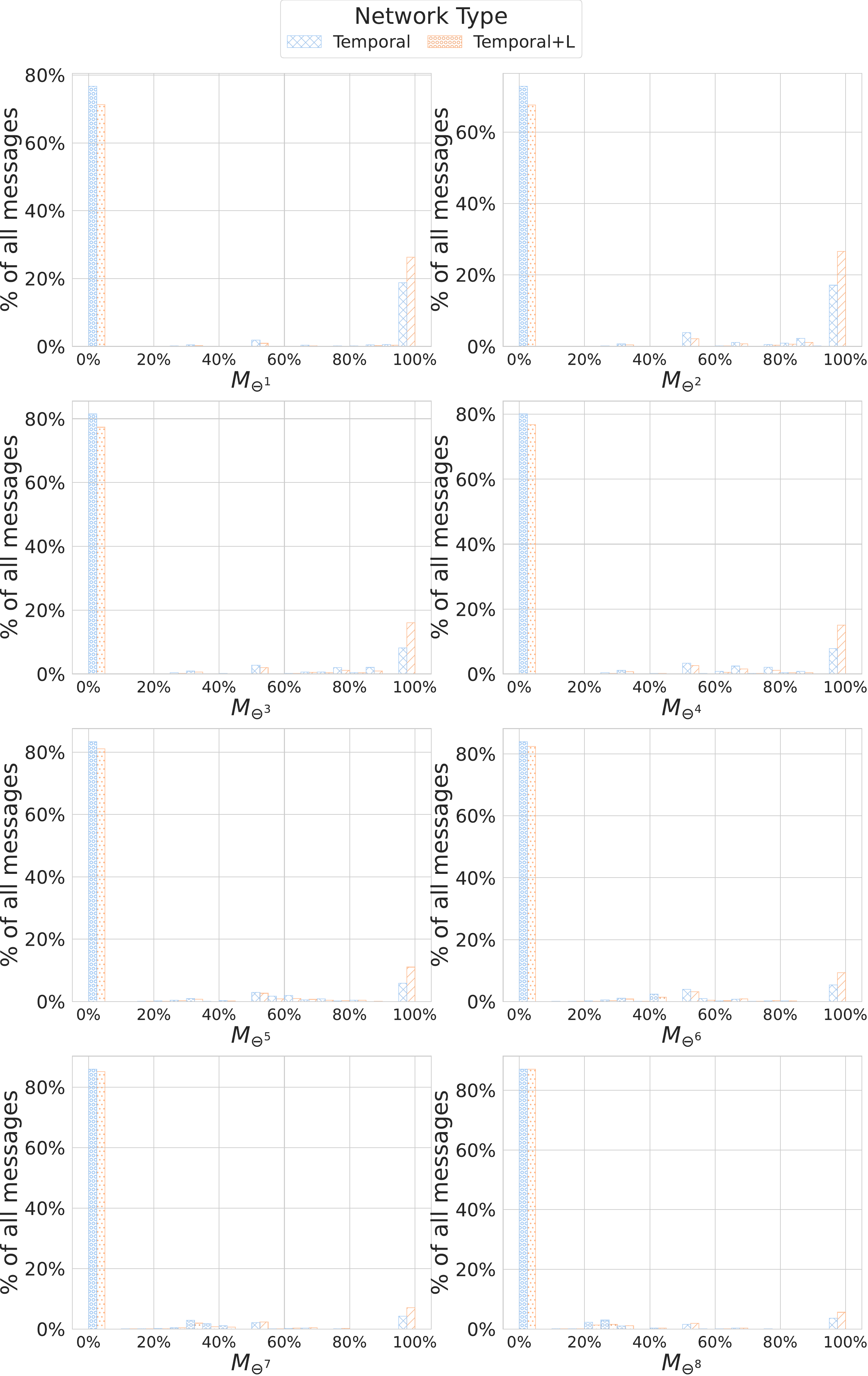}
	\caption{Usage of messages compared to their $M_{\ominus^h}$ value for the \textit{Temporal} agents.}
	\label{fig: histogram h8 temporal}
\end{figure}

\begin{figure}[h]
	\centering
	\includegraphics[width=0.70\textwidth]{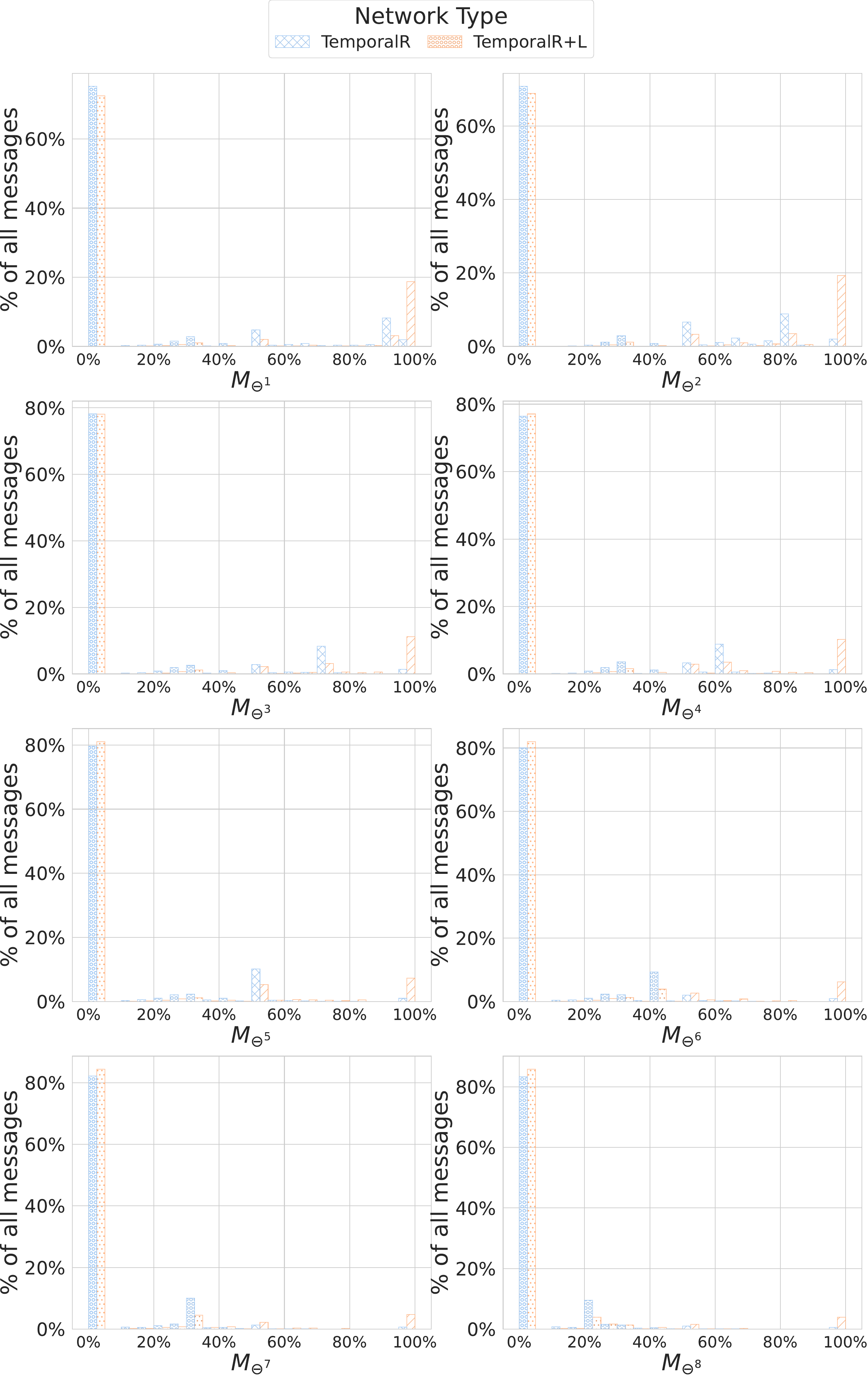}
	\caption{Usage of messages compared to their $M_{\ominus^h}$ value for the \textit{TemporalR} agents.}
	\label{fig: histogram h8 temporalr}
\end{figure}

\clearpage

\subsection{Analysis within Network Types}

In this section, the detailed values of all statistical significance tests are provided.

\subsubsection{Normality Tests}

\begin{table}[h]
	\centering
	\begin{adjustbox}{width=\textwidth}
		\begin{tabular}{llllllll}
			Network Type & Training Env & Always Same & Never Same & RG   & RG Hard & TRG  & TRG Hard \\
			\toprule
			Base         & RG           & 0.01        & 0.01       & 0.01 & 0.02    & 0.01 & 0.02     \\
			             & TRG          & 0.39        & 0.04       & 0.02 & 0.01    & 0.01 & 0.01     \\
			Base+L       & RG           & 0.01        & 0.01       & 0.01 & 0.73    & 0.08 & 0.01     \\
			             & TRG          & 0.18        & 0.25       & 0.98 & 0.42    & 0.24 & 0.01     \\
			\midrule
			Temporal     & RG           & 0.11        & 0.65       & 0.22 & 0.86    & 0.69 & 0.01     \\
			             & TRG          & 0.01        & 0.01       & 0.01 & 0.02    & 0.01 & 0.01     \\
			Temporal+L   & RG           & 0.02        & 0.07       & 0.08 & 0.46    & 0.48 & 0.01     \\
			             & TRG          & 0.03        & 0.78       & 0.82 & 0.80    & 0.60 & 0.01     \\
			\midrule
			TemporalR    & RG           & 0.09        & 0.68       & 0.27 & 0.04    & 0.16 & 0.01     \\
			             & TRG          & 0.02        & 0.01       & 0.02 & 0.05    & 0.25 & 0.01     \\
			TemporalR+L  & RG           & 0.03        & 0.01       & 0.37 & 0.28    & 0.63 & 0.01     \\
			             & TRG          & 0.06        & 0.10       & 0.21 & 0.30    & 0.34 & 0.01     \\
		\end{tabular}
	\end{adjustbox}
	\caption{$p$ values for the normality test on the accuracy scores.}
	\label{tab: normality acc}
\end{table}

\begin{table}[h]
	\centering
	\begin{adjustbox}{width=\textwidth}
		\begin{tabular}{llllllll}
			Network Type & Training Env & Always Same & Never Same & RG  & RG Hard & TRG & TRG Hard \\
			\toprule
			Base         & RG           & NaN         & N/A        & NaN & NaN     & 0   & 0        \\
			             & TRG          & NaN         & N/A        & NaN & NaN     & 0   & 0        \\
			Base+L       & RG           & NaN         & N/A        & NaN & NaN     & 0   & 0        \\
			             & TRG          & NaN         & N/A        & NaN & NaN     & 0   & 0        \\
			\midrule
			Temporal     & RG           & 0           & N/A        & NaN & NaN     & 0   & 0        \\
			             & TRG          & 0           & N/A        & NaN & NaN     & 0   & 0        \\
			Temporal+L   & RG           & 0           & N/A        & NaN & NaN     & 0   & 0        \\
			             & TRG          & 0           & N/A        & NaN & NaN     & 0   & 0        \\
			\midrule
			TemporalR    & RG           & 0           & N/A        & NaN & NaN     & 0   & 0        \\
			             & TRG          & 0           & N/A        & NaN & NaN     & 0   & 0        \\
			TemporalR+L  & RG           & 0           & N/A        & NaN & NaN     & 0   & 0        \\
			             & TRG          & 0           & N/A        & NaN & NaN     & 0   & 0        \\ 
		\end{tabular}
	\end{adjustbox}
	\caption{$p$ values for the normality test on the $M_{\ominus^n}$ scores.}
	\label{tab: normality temporality}
\end{table}

\begin{table}[h]
	\centering
	\begin{tabular}{lllll}
		Network Type & Training Env & TopSim & Posdis & Bosdis \\
		\toprule
		Base         & RG           & 0.01   & 0.01   & 0.75   \\
		             & TRG          & 0.44   & 0.01   & 0.90   \\
		Base+L       & RG           & 0.31   & 0.02   & 0.05   \\
		             & TRG          & 0.91   & 0.01   & 0.01   \\
		\midrule
		Temporal     & RG           & 0.11   & 0.01   & 0.02   \\
		             & TRG          & 0.38   & 0.02   & 0.01   \\
		Temporal+L   & RG           & 0.05   & 0.01   & 0.14   \\
		             & TRG          & 0.38   & 0.01   & 0.80   \\
		\midrule
		TemporalR    & RG           & 0.97   & 0.10   & 0.32   \\
		             & TRG          & 0.34   & 0.08   & 0.52   \\
		TemporalR+L  & RG           & 0.02   & 0.01   & 0.83   \\
		             & TRG          & 0.01   & 0.02   & 0.01   \\ 
	\end{tabular}
	\caption{$p$ values for the normality test on the compositionality scores.}
	\label{tab: normality comp}
\end{table}

\clearpage

\subsubsection{Accuracy Kruskal-Wallis H-test P-Values within Network Types}

\begin{table}[h]
	\centering
	\begin{adjustbox}{width=\textwidth}
		\begin{tabular}{lrrrrrr}
			\toprule
			          & Always Same & Never Same & RG       & RG Hard & TRG      & TRG Hard \\
			\midrule
			Base      & 0           & 0          & 0        & 0       & 0        & 0.060617 \\
			Temporal  & 0           & 0          & 0        & 0       & 0        & 0.000197 \\
			TemporalR & 0           & 0          & 0        & 0       & 0        & 0.004879 \\
			\bottomrule
		\end{tabular}
	\end{adjustbox}
	\caption{$p$ values of the Kruskal-Wallis H-test for the accuracy scores within a given network type.}
	\label{tab: acc kw pval within}
\end{table}

\subsubsection{Accuracy Post-Hoc Conover Analysis within Network Types}

\begin{table}[h]
	\centering
	\begin{tabular}{lrrrr}
		\toprule
		           & Base+L RG & Base RG  & Base+L TRG & Base RG  \\
		\midrule
		Base+L RG  & 1         & 0.433259 & 0          & 0.498274 \\
		Base RG    & 0.433259  & 1        & 0          & 0.171310 \\
		Base+L TRG & 0         & 0        & 1          & 0        \\
		Base RG    & 0.498274  & 0.171310 & 0          & 1        \\
		\bottomrule
	\end{tabular}
	\caption{$p$ values of the post-hoc Conover test for the accuracy scores on the Always Same environment.}
	\label{tab: acc phc pval within as base}
\end{table}

\begin{table}[h]
	\centering
	\begin{tabular}{lrrrr}
		\toprule
		           & Base+L RG & Base RG  & Base+L TRG & Base RG  \\
		\midrule
		Base+L RG  & 1         & 0.433899 & 0          & 0.582182 \\
		Base RG    & 0.433899  & 1        & 0          & 0.226648 \\
		Base+L TRG & 0         & 0        & 1          & 0        \\
		Base RG    & 0.582182  & 0.226648 & 0          & 1        \\
		\bottomrule
	\end{tabular}
	\caption{$p$ values of the post-hoc Conover test for the accuracy scores on the Never Same environment.}
	\label{tab: acc phc pval within ns base}
\end{table}

\begin{table}[h]
	\centering
	\begin{tabular}{lrrrr}
		\toprule
		           & Base+L RG & Base RG  & Base+L TRG & Base RG  \\
		\midrule
		Base+L RG  & 1         & 0.149439 & 0          & 0.668939 \\
		Base RG    & 0.149439  & 1        & 0          & 0.083620 \\
		Base+L TRG & 0         & 0        & 1          & 0        \\
		Base RG    & 0.668939  & 0.083620 & 0          & 1        \\
		\bottomrule
	\end{tabular}
	\caption{$p$ values of the post-hoc Conover test for the accuracy scores on the RG environment.}
	\label{tab: acc phc pval within rg base}
\end{table}

\begin{table}[h]
	\centering
	\begin{tabular}{lrrrr}
		\toprule
		           & Base+L RG & Base RG  & Base+L TRG & Base RG  \\
		\midrule
		Base+L RG  & 1         & 0.346337 & 0          & 0.346337 \\
		Base RG    & 0.346337  & 1        & 0          & 0.040696 \\
		Base+L TRG & 0         & 0        & 1          & 0        \\
		Base RG    & 0.346337  & 0.040696 & 0          & 1        \\
		\bottomrule
	\end{tabular}
	\caption{$p$ values of the post-hoc Conover test for the accuracy scores on the RG Hard environment.}
	\label{tab: acc phc pval within rgh base}
\end{table}

\begin{table}[h]
	\centering
	\begin{tabular}{lrrrr}
		\toprule
		           & Base+L RG & Base RG  & Base+L TRG & Base RG  \\
		\midrule
		Base+L RG  & 1         & 0.185363 & 0          & 0.185363 \\
		Base RG    & 0.185363  & 1        & 0          & 0.005564 \\
		Base+L TRG & 0         & 0        & 1          & 0        \\
		Base RG    & 0.185363  & 0.005564 & 0          & 1        \\
		\bottomrule
	\end{tabular}
	\caption{$p$ values of the post-hoc Conover test for the accuracy scores on the TRG environment.}
	\label{tab: acc phc pval within trg base}
\end{table}

\begin{table}[h]
	\centering
	\begin{tabular}{lrrrr}
		\toprule
		           & Base+L RG & Base RG  & Base+L TRG & Base RG  \\
		\midrule
		Base+L RG  & 1         & 1        & 0.175554   & 1        \\
		Base RG    & 1         & 1        & 0.087409   & 1        \\
		Base+L TRG & 0.175554  & 0.087409 & 1          & 0.175554 \\
		Base RG    & 1         & 1        & 0.175554   & 1        \\
		\bottomrule
	\end{tabular}
	\caption{$p$ values of the post-hoc Conover test for the accuracy scores on the TRG Hard environment.}
	\label{tab: acc phc pval within trgh base}
\end{table}

\begin{table}[h]
	\centering
	\begin{adjustbox}{width=\textwidth}
		\begin{tabular}{lrrrr}
			\toprule
			               & Temporal+L RG & Temporal RG & Temporal+L TRG & Temporal RG \\
			\midrule
			Temporal+L RG  & 1             & 1           & 0              & 1           \\
			Temporal RG    & 1             & 1           & 0              & 1           \\
			Temporal+L TRG & 0             & 0           & 1              & 0           \\
			Temporal RG    & 1             & 1           & 0              & 1           \\
			\bottomrule
		\end{tabular}
	\end{adjustbox}
	\caption{$p$ values of the post-hoc Conover test for the accuracy scores on the Always Same environment.}
\end{table}

\begin{table}[h]
	\centering
	\begin{adjustbox}{width=\textwidth}
		\begin{tabular}{lrrrr}
			\toprule
			               & Temporal+L RG & Temporal RG & Temporal+L TRG & Temporal RG \\
			\midrule
			Temporal+L RG  & 1             & 0.876314    & 0              & 0.114027    \\
			Temporal RG    & 0.876314      & 1           & 0              & 0.114027    \\
			Temporal+L TRG & 0             & 0           & 1              & 0           \\
			Temporal RG    & 0.114027      & 0.114027    & 0              & 1           \\
			\bottomrule
		\end{tabular}
	\end{adjustbox}
	\caption{$p$ values of the post-hoc Conover test for the accuracy scores on the Never Same environment.}
	\label{tab: acc phc pval within ns temporal}
\end{table}

\begin{table}[h]
	\centering
	\begin{adjustbox}{width=\textwidth}
		\begin{tabular}{lrrrr}
			\toprule
			               & Temporal+L RG & Temporal RG & Temporal+L TRG & Temporal RG \\
			\midrule
			Temporal+L RG  & 1             & 0.384963    & 0              & 0.635388    \\
			Temporal RG    & 0.384963      & 1           & 0              & 0.229984    \\
			Temporal+L TRG & 0             & 0           & 1              & 0           \\
			Temporal RG    & 0.635388      & 0.229984    & 0              & 1           \\
			\bottomrule
		\end{tabular}
	\end{adjustbox}
	\caption{$p$ values of the post-hoc Conover test for the accuracy scores on the RG environment.}
	\label{tab: acc phc pval within rg temporal}
\end{table}

\begin{table}[h]
	\centering
	\begin{adjustbox}{width=\textwidth}
		\begin{tabular}{lrrrr}
			\toprule
			               & Temporal+L RG & Temporal RG & Temporal+L TRG & Temporal RG \\
			\midrule
			Temporal+L RG  & 1             & 0.056940    & 0              & 0.000033    \\
			Temporal RG    & 0.056940      & 1           & 0              & 0.017182    \\
			Temporal+L TRG & 0             & 0           & 1              & 0           \\
			Temporal RG    & 0.000033      & 0.017182    & 0              & 1           \\
			\bottomrule
		\end{tabular}
	\end{adjustbox}
	\caption{$p$ values of the post-hoc Conover test for the accuracy scores on the RG Hard environment.}
	\label{tab: acc phc pval within rgh temporal}
\end{table}

\begin{table}[h]
	\centering
	\begin{adjustbox}{width=\textwidth}
		\begin{tabular}{lrrrr}
			\toprule
			               & Temporal+L RG & Temporal RG & Temporal+L TRG & Temporal RG \\
			\midrule
			Temporal+L RG  & 1             & 1           & 0              & 1           \\
			Temporal RG    & 1             & 1           & 0              & 1           \\
			Temporal+L TRG & 0             & 0           & 1              & 0           \\
			Temporal RG    & 1             & 1           & 0              & 1           \\
			\bottomrule
		\end{tabular}
	\end{adjustbox}
	\caption{$p$ values of the post-hoc Conover test for the accuracy scores on the TRG environment.}
	\label{tab: acc phc pval within trg temporal}
\end{table}

\begin{table}[h]
	\centering
	\begin{adjustbox}{width=\textwidth}
		\begin{tabular}{lrrrr}
			\toprule
			               & Temporal+L RG & Temporal RG & Temporal+L TRG & Temporal RG \\
			\midrule
			Temporal+L RG  & 1             & 0.993318    & 0.000135       & 0.691546    \\
			Temporal RG    & 0.993318      & 1           & 0.001456       & 0.993318    \\
			Temporal+L TRG & 0.000135      & 0.001456    & 1              & 0.006804    \\
			Temporal RG    & 0.691546      & 0.993318    & 0.006804       & 1           \\
			\bottomrule
		\end{tabular}
	\end{adjustbox}
	\caption{$p$ values of the post-hoc Conover test for the accuracy scores on the TRG Hard environment.}
	\label{tab: acc phc pval within trgh temporal}
\end{table}

\begin{table}[h]
	\centering
	\begin{adjustbox}{width=\textwidth}
		\begin{tabular}{lrrrr}
			\toprule
			                & TemporalR+L RG & TemporalR RG & TemporalR+L TRG & TemporalR RG \\
			\midrule
			TemporalR+L RG  & 1              & 0.442920     & 0               & 0.442920     \\
			TemporalR RG    & 0.442920       & 1            & 0               & 0.071027     \\
			TemporalR+L TRG & 0              & 0            & 1               & 0            \\
			TemporalR RG    & 0.442920       & 0.071027     & 0               & 1            \\
			\bottomrule
		\end{tabular}
	\end{adjustbox}
	\caption{$p$ values of the post-hoc Conover test for the accuracy scores on the Always Same environment.}
\end{table}

\begin{table}[h]
	\centering
	\begin{adjustbox}{width=\textwidth}
		\begin{tabular}{lrrrr}
			\toprule
			                & TemporalR+L RG & TemporalR RG & TemporalR+L TRG & TemporalR RG \\
			\midrule
			TemporalR+L RG  & 1              & 0.224358     & 0               & 0.084598     \\
			TemporalR RG    & 0.224358       & 1            & 0               & 0.004183     \\
			TemporalR+L TRG & 0              & 0            & 1               & 0            \\
			TemporalR RG    & 0.084598       & 0.004183     & 0               & 1            \\
			\bottomrule
		\end{tabular}
	\end{adjustbox}
	\caption{$p$ values of the post-hoc Conover test for the accuracy scores on the Never Same environment.}
		
\end{table}

\begin{table}[h]
	\centering
	\begin{adjustbox}{width=\textwidth}
		\begin{tabular}{lcccc}
			\toprule
			                & TemporalR+L RG & TemporalR RG & TemporalR+L TRG & TemporalR RG \\
			\midrule
			TemporalR+L RG  & 1              & 0.009688     & 0               & 0.512991     \\
			TemporalR RG    & 0.009688       & 1            & 0               & 0.040779     \\
			TemporalR+L TRG & 0              & 0            & 1               & 0            \\
			TemporalR RG    & 0.512991       & 0.040779     & 0               & 1            \\
			\bottomrule
		\end{tabular}
	\end{adjustbox}
	\caption{$p$ values of the post-hoc Conover test for the accuracy scores on the RG environment.}
		
\end{table}

\begin{table}[h]
	\centering
	\begin{adjustbox}{width=\textwidth}
		\begin{tabular}{lrrrr}
			\toprule
			                & TemporalR+L RG & TemporalR RG & TemporalR+L TRG & TemporalR RG \\
			\midrule
			TemporalR+L RG  & 1              & 0.589040     & 0               & 0.871791     \\
			TemporalR RG    & 0.589040       & 1            & 0               & 0.589040     \\
			TemporalR+L TRG & 0              & 0            & 1               & 0            \\
			TemporalR RG    & 0.871791       & 0.589040     & 0               & 1            \\
			\bottomrule
		\end{tabular}
	\end{adjustbox}
	\caption{$p$ values of the post-hoc Conover test for the accuracy scores on the RG Hard environment.}
		
\end{table}

\begin{table}[h]
	\centering
	\begin{adjustbox}{width=\textwidth}
		\begin{tabular}{lrrrr}
			\toprule
			                & TemporalR+L RG & TemporalR RG & TemporalR+L TRG & TemporalR RG \\
			\midrule
			TemporalR+L RG  & 1              & 0.015392     & 0               & 0.481321     \\
			TemporalR RG    & 0.015392       & 1            & 0               & 0.002605     \\
			TemporalR+L TRG & 0              & 0            & 1               & 0            \\
			TemporalR RG    & 0.481321       & 0.002605     & 0               & 1            \\
			\bottomrule
		\end{tabular}
	\end{adjustbox}
	\caption{$p$ values of the post-hoc Conover test for the accuracy scores on the TRG environment.}
\end{table}

\begin{table}[h]
	\centering
	\begin{adjustbox}{width=\textwidth}
		\begin{tabular}{lrrrr}
			\toprule
			                & TemporalR+L RG & TemporalR RG & TemporalR+L TRG & TemporalR RG \\
			\midrule
			TemporalR+L RG  & 1              & 1            & 0.011330        & 1            \\
			TemporalR RG    & 1              & 1            & 0.012271        & 1            \\
			TemporalR+L TRG & 0.011330       & 0.012271     & 1               & 0.018447     \\
			TemporalR RG    & 1              & 1            & 0.018447        & 1            \\
			\bottomrule
		\end{tabular}
	\end{adjustbox}
	\caption{$p$ values of the post-hoc Conover test for the accuracy scores on the TRG Hard environment.}
		
\end{table}

\clearpage

\subsubsection{Topographic Similarity Post-Hoc Conover Analysis within Network Types}

\begin{table}[h]
	\centering
	 
	\begin{tabular}{lrrrr}
		\toprule
		           & Base+L RG & Base RG & Base+L TRG & Base RG \\
		\midrule
		Base+L RG  & 1         & 1       & 0          & 1       \\
		Base RG    & 1         & 1       & 0          & 1       \\
		Base+L TRG & 0         & 0       & 1          & 0       \\
		Base RG    & 1         & 1       & 0          & 1       \\
		\bottomrule
	\end{tabular}
	\caption{$p$ values of the post-hoc Conover test for the topographic similarity scores.}
		
\end{table}

\begin{table}[h]
	\centering
	\begin{adjustbox}{width=\textwidth}
		\begin{tabular}{lrrrr}
			\toprule
			               & Temporal+L RG & Temporal RG & Temporal+L TRG & Temporal RG \\
			\midrule
			Temporal+L RG  & 1             & 0.271644    & 0              & 0.075936    \\
			Temporal RG    & 0.271644      & 1           & 0              & 0.454652    \\
			Temporal+L TRG & 0             & 0           & 1              & 0           \\
			Temporal RG    & 0.075936      & 0.454652    & 0              & 1           \\
			\bottomrule
		\end{tabular}
	\end{adjustbox}
	\caption{$p$ values of the post-hoc Conover test for the topographic similarity scores.}
		
\end{table}

\begin{table}[h]
	\centering
	\begin{adjustbox}{width=\textwidth}
		\begin{tabular}{lrrrr}
			\toprule
			                & TemporalR+L RG & TemporalR RG & TemporalR+L TRG & TemporalR RG \\
			\midrule
			TemporalR+L RG  & 1              & 0.000081     & 0               & 0            \\
			TemporalR RG    & 0.000081       & 1            & 0               & 0.131500     \\
			TemporalR+L TRG & 0              & 0            & 1               & 0            \\
			TemporalR RG    & 0              & 0.131500     & 0               & 1            \\
			\bottomrule
		\end{tabular}
	\end{adjustbox}
	\caption{$p$ values of the post-hoc Conover test for the topographic similarity scores.}
		
\end{table}

\subsubsection{Positional Disentanglement Post-Hoc Conover Analysis within Network Types}

\begin{table}[h]
	\centering
	\begin{tabular}{lrrrr}
		\toprule
		           & Base+L RG & Base RG & Base+L TRG & Base RG  \\
		\midrule
		Base+L RG  & 1         & 1       & 0          & 0.967127 \\
		Base RG    & 1         & 1       & 0          & 1        \\
		Base+L TRG & 0         & 0       & 1          & 0        \\
		Base RG    & 0.967127  & 1       & 0          & 1        \\
		\bottomrule
	\end{tabular}
	\caption{$p$ values of the post-hoc Conover test for the posdis scores.}
\end{table}

\begin{table}[h]
	\centering
	\begin{adjustbox}{width=\textwidth}
		\begin{tabular}{lrrrr}
			\toprule
			               & Temporal+L RG & Temporal RG & Temporal+L TRG & Temporal RG \\
			\midrule
			Temporal+L RG  & 1             & 1           & 1              & 0.082919    \\
			Temporal RG    & 1             & 1           & 1              & 0.015836    \\
			Temporal+L TRG & 1             & 1           & 1              & 0.045470    \\
			Temporal RG    & 0.082919      & 0.015836    & 0.045470       & 1           \\
			\bottomrule
		\end{tabular}
	\end{adjustbox}
	\caption{$p$ values of the post-hoc Conover test for the posdis scores.}
\end{table}

\begin{table}[h]
	\centering
	\begin{adjustbox}{width=\textwidth}
		\begin{tabular}{lrrrr}
			\toprule
			                & TemporalR+L RG & TemporalR RG & TemporalR+L TRG & TemporalR RG \\
			\midrule
			TemporalR+L RG  & 1              & 1            & 0.149973        & 1            \\
			TemporalR RG    & 1              & 1            & 0.042950        & 1            \\
			TemporalR+L TRG & 0.149973       & 0.042950     & 1               & 0.248096     \\
			TemporalR RG    & 1              & 1            & 0.248096        & 1            \\
			\bottomrule
		\end{tabular}
	\end{adjustbox}
	\caption{$p$ values of the post-hoc Conover test for the posdis scores.}
\end{table}

\clearpage

\subsubsection{Bag-of-Words Disentanglement Post-Hoc Conover Analysis within Network Types}

\begin{table}[h]
	\centering
		\begin{tabular}{lrrrr}
			\toprule
			           & Base+L RG & Base RG  & Base+L TRG & Base RG  \\
			\midrule
			Base+L RG  & 1         & 1        & 0.001331   & 0.843365 \\
			Base RG    & 1         & 1        & 0.008259   & 1        \\
			Base+L TRG & 0.001331  & 0.008259 & 1          & 0.034571 \\
			Base RG    & 0.843365  & 1        & 0.034571   & 1        \\
			\bottomrule
		\end{tabular}
	\caption{$p$ values of the post-hoc Conover test for the bosdis scores.}
\end{table}

\begin{table}[h]
	\centering
	\begin{adjustbox}{width=\textwidth}
		\begin{tabular}{lrrrr}
			\toprule
			               & Temporal+L RG & Temporal RG & Temporal+L TRG & Temporal RG \\
			\midrule
			Temporal+L RG  & 1             & 0.023856    & 0              & 0.000197    \\
			Temporal RG    & 0.023856      & 1           & 0              & 0.135339    \\
			Temporal+L TRG & 0             & 0           & 1              & 0           \\
			Temporal RG    & 0.000197      & 0.135339    & 0              & 1           \\
			\bottomrule
		\end{tabular}
	\end{adjustbox}
	\caption{$p$ values of the post-hoc Conover test for the bosdis scores.}
\end{table}

\begin{table}[h]
	\centering
	\begin{adjustbox}{width=\textwidth}
		\begin{tabular}{lrrrr}
			\toprule
			                & TemporalR+L RG & TemporalR RG & TemporalR+L TRG & TemporalR RG \\
			\midrule
			TemporalR+L RG  & 1              & 0.043151     & 0               & 0.878130     \\
			TemporalR RG    & 0.043151       & 1            & 0               & 0.043455     \\
			TemporalR+L TRG & 0              & 0            & 1               & 0            \\
			TemporalR RG    & 0.878130       & 0.043455     & 0               & 1            \\
			\bottomrule
		\end{tabular}
	\end{adjustbox}
	\caption{$p$ values of the post-hoc Conover test for the bosdis scores.}
\end{table}

\clearpage

\subsection{Analysis Between Network Types}
\subsubsection{Accuracy Post-Hoc Conover Analysis across Network Types}

We do not list all the Kruskal-Wallis H-test $p$ values individually, as they are all below the threshold value of 0.001.

\begin{table}[h]
	\centering
	\begin{tabular}{lrrrr}
		\toprule
		          & Base     & Temporal & TemporalR \\
		\midrule
		Base      & 1        & 0.000317 & 0.330549  \\
		Temporal  & 0.000317 & 1        & 0.006967  \\
		TemporalR & 0.330549 & 0.006967 & 1         \\
		\bottomrule
	\end{tabular}
	\caption{$p$ values of the post-hoc Conover test for the accuracy scores on the Always Same environment.}
	\label{tab: acc phc pval across as}
\end{table}

\begin{table}[h]
	\centering
	\begin{tabular}{lrrrr}
		\toprule
		          & Base     & Temporal & TemporalR \\
		\midrule
		Base      & 1        & 0.000005 & 0.235654  \\
		Temporal  & 0.000005 & 1        & 0.000513  \\
		TemporalR & 0.235654 & 0.000513 & 1         \\
		\bottomrule
	\end{tabular}
	\caption{$p$ values of the post-hoc Conover test for the accuracy scores on the Never Same environment.}
	\label{tab: acc phc pval across ns}
\end{table}

\begin{table}[h]
	\centering
	\begin{tabular}{lrrrr}
		\toprule
		          & Base     & Temporal & TemporalR \\
		\midrule
		Base      & 1        & 0.000001 & 0.213130  \\
		Temporal  & 0.000001 & 1        & 0.000186  \\
		TemporalR & 0.213130 & 0.000186 & 1         \\
		\bottomrule
	\end{tabular}
	\caption{$p$ values of the post-hoc Conover test for the accuracy scores on the RG environment.}
	\label{tab: acc phc pval across rg}
\end{table}

\begin{table}[h]
	\centering
	\begin{tabular}{lrrrr}
		\toprule
		          & Base     & Temporal & TemporalR \\
		\midrule
		Base      & 1        & 0.000184 & 0.909507  \\
		Temporal  & 0.000184 & 1        & 0.000195  \\
		TemporalR & 0.909507 & 0.000195 & 1         \\
		\bottomrule
	\end{tabular}
	\caption{$p$ values of the post-hoc Conover test for the accuracy scores on the RG Hard environment.}
	\label{tab: acc phc pval across rgh}
\end{table}

\begin{table}[h]
	\centering
	\begin{tabular}{lrrrr}
		\toprule
		          & Base     & Temporal & TemporalR \\
		\midrule
		Base      & 1        & 0.000011 & 0.337959  \\
		Temporal  & 0.000011 & 1        & 0.000422  \\
		TemporalR & 0.337959 & 0.000422 & 1         \\
		\bottomrule
	\end{tabular}
	\caption{$p$ values of the post-hoc Conover test for the accuracy scores on the TRG environment.}
	\label{tab: acc phc pval across trg}
\end{table}

\begin{table}[h]
	\centering
	\begin{tabular}{lrrrr}
		\toprule
		          & Base     & Temporal & TemporalR \\
		\midrule
		Base      & 1        & 0.676719 & 0.763523  \\
		Temporal  & 0.676719 & 1        & 0.724056  \\
		TemporalR & 0.763523 & 0.724056 & 1         \\
		\bottomrule
	\end{tabular}
	\caption{$p$ values of the post-hoc Conover test for the accuracy scores on the TRG Hard environment.}
	\label{tab: acc phc pval across trgh}
\end{table}

\clearpage

\subsubsection{Compositionality Post-Hoc Conover Analysis across Network Types}

We provide a full breakdown of the post-hoc Conover analysis of the compositionality scores. However, the $p$ value of the Kruskal-Wallis H-test for the posdis metric was above the 0.001 threshold ($p=0.10$), and so this analysis may not be statistically significant.

\begin{table}[h]
	\centering
	\begin{tabular}{lrrrr}
		\toprule
		          & Base     & Temporal & TemporalR \\
		\midrule
		Base      & 1        & 0        & 0.000020  \\
		Temporal  & 0        & 1        & 0.001046  \\
		TemporalR & 0.000020 & 0.001046 & 1         \\
		\bottomrule
	\end{tabular}
	\caption{$p$ values of the post-hoc Conover test for the topographic similarity scores.}
	\label{tab: topsim pval}
\end{table}

\begin{table}[h]
	\centering
	\begin{tabular}{lrrrr}
		\toprule
		          & Base     & Temporal & TemporalR \\
		\midrule
		Base      & 1        & 0.274577 & 1         \\
		Temporal  & 0.274577 & 1        & 0.294687  \\
		TemporalR & 1        & 0.294687 & 1         \\
		\bottomrule
	\end{tabular}
	\caption{$p$ values of the post-hoc Conover test for the posdis scores.}
	\label{tab: posdis pval}
\end{table}

\begin{table}[h]
	\centering
	\begin{tabular}{lrrrr}
		\toprule
		          & Base     & Temporal & TemporalR \\
		\midrule
		Base      & 1        & 0.003442 & 0.813492  \\
		Temporal  & 0.003442 & 1        & 0.032353  \\
		TemporalR & 0.813492 & 0.032353 & 1         \\
		\bottomrule
	\end{tabular}
	\caption{$p$ values of the post-hoc Conover test for the bosdis scores.}
	\label{tab: bosdis pval}
\end{table}

\clearpage

\subsubsection{Temporality Post-Hoc Conover Analysis across Network Types}

We do not list all the Kruskal-Wallis H-test $p$ values individually, as they are all below the threshold value of 0.001.

\begin{table}[h]
	\centering
	\begin{tabular}{lrrrr}
		\toprule
		          & Base & Temporal & TemporalR \\
		\midrule
		Base      & 1    & 0        & 0         \\
		Temporal  & 0    & 1        & 0         \\
		TemporalR & 0    & 0        & 1         \\
		\bottomrule
	\end{tabular}
	\caption{$p$ values of the post-hoc Conover test for the $M_{\ominus^1}$ scores on the Always Same environment.}
		
\end{table}

\begin{table}[h]
	\centering
	\begin{tabular}{lrrrr}
		\toprule
		          & Base & Temporal & TemporalR \\
		\midrule
		Base      & 1    & 0        & 0         \\
		Temporal  & 0    & 1        & 0.049232  \\
		TemporalR & 0    & 0.049232 & 1         \\
		\bottomrule
	\end{tabular}
	\caption{$p$ values of the post-hoc Conover test for the $M_{\ominus^2}$ scores on the Always Same environment.}
		
\end{table}

\begin{table}[h]
	\centering
	\begin{tabular}{lrrrr}
		\toprule
		          & Base     & Temporal & TemporalR \\
		\midrule
		Base      & 1        & 0        & 0.000001  \\
		Temporal  & 0        & 1        & 0         \\
		TemporalR & 0.000001 & 0        & 1         \\
		\bottomrule
	\end{tabular}
	\caption{$p$ values of the post-hoc Conover test for the $M_{\ominus^3}$ scores on the Always Same environment.}
		
\end{table}

\begin{table}[h]
	\centering
	\begin{tabular}{lrrrr}
		\toprule
		          & Base & Temporal & TemporalR \\
		\midrule
		Base      & 1    & 0        & 0         \\
		Temporal  & 0    & 1        & 0         \\
		TemporalR & 0    & 0        & 1         \\
		\bottomrule
	\end{tabular}
	\caption{$p$ values of the post-hoc Conover test for the $M_{\ominus^4}$ scores on the Always Same environment.}
		
\end{table}

\begin{table}[h]
	\centering
	\begin{tabular}{lrrrr}
		\toprule
		          & Base & Temporal & TemporalR \\
		\midrule
		Base      & 1    & 0        & 0         \\
		Temporal  & 0    & 1        & 0         \\
		TemporalR & 0    & 0        & 1         \\
		\bottomrule
	\end{tabular}
	\caption{$p$ values of the post-hoc Conover test for the $M_{\ominus^5}$ scores on the Always Same environment.}
		
\end{table}

\begin{table}[h]
	\centering
	\begin{tabular}{lrrrr}
		\toprule
		          & Base & Temporal & TemporalR \\
		\midrule
		Base      & 1    & 0        & 0         \\
		Temporal  & 0    & 1        & 0         \\
		TemporalR & 0    & 0        & 1         \\
		\bottomrule
	\end{tabular}
	\caption{$p$ values of the post-hoc Conover test for the $M_{\ominus^6}$ scores on the Always Same environment.}
		
\end{table}

\begin{table}[h]
	\centering
	\begin{tabular}{lrrrr}
		\toprule
		          & Base & Temporal & TemporalR \\
		\midrule
		Base      & 1    & 0        & 0         \\
		Temporal  & 0    & 1        & 0         \\
		TemporalR & 0    & 0        & 1         \\
		\bottomrule
	\end{tabular}
	\caption{$p$ values of the post-hoc Conover test for the $M_{\ominus^7}$ scores on the Always Same environment.}
		
\end{table}

\begin{table}[h]
	\centering
	\begin{tabular}{lrrrr}
		\toprule
		          & Base & Temporal & TemporalR \\
		\midrule
		Base      & 1    & 0        & 0         \\
		Temporal  & 0    & 1        & 0         \\
		TemporalR & 0    & 0        & 1         \\
		\bottomrule
	\end{tabular}
	\caption{$p$ values of the post-hoc Conover test for the $M_{\ominus^8}$ scores on the Always Same environment.}
		
\end{table}

\begin{table}[h]
	\centering
	\begin{tabular}{lrrrr}
		\toprule
		          & Base & Temporal & TemporalR \\
		\midrule
		Base      & 1    & 0        & 0         \\
		Temporal  & 0    & 1        & 0.031283  \\
		TemporalR & 0    & 0.031283 & 1         \\
		\bottomrule
	\end{tabular}
	\caption{$p$ values of the post-hoc Conover test for the $M_{\ominus^1}$ scores on the TRG environment.}
		
\end{table}

\begin{table}[h]
	\centering
	\begin{tabular}{lrrrr}
		\toprule
		          & Base & Temporal & TemporalR \\
		\midrule
		Base      & 1    & 0        & 0         \\
		Temporal  & 0    & 1        & 0         \\
		TemporalR & 0    & 0        & 1         \\
		\bottomrule
	\end{tabular}
	\caption{$p$ values of the post-hoc Conover test for the $M_{\ominus^2}$ scores on the TRG environment.}
		
\end{table}

\begin{table}[h]
	\centering
	\begin{tabular}{lrrrr}
		\toprule
		          & Base & Temporal & TemporalR \\
		\midrule
		Base      & 1    & 0        & 0         \\
		Temporal  & 0    & 1        & 0         \\
		TemporalR & 0    & 0        & 1         \\
		\bottomrule
	\end{tabular}
	\caption{$p$ values of the post-hoc Conover test for the $M_{\ominus^3}$ scores on the TRG environment.}
		
\end{table}

\begin{table}[h]
	\centering
	\begin{tabular}{lrrrr}
		\toprule
		          & Base & Temporal & TemporalR \\
		\midrule
		Base      & 1    & 0        & 0         \\
		Temporal  & 0    & 1        & 0         \\
		TemporalR & 0    & 0        & 1         \\
		\bottomrule
	\end{tabular}
	\caption{$p$ values of the post-hoc Conover test for the $M_{\ominus^4}$ scores on the TRG environment.}
		
\end{table}

\begin{table}[h]
	\centering
	\begin{tabular}{lrrrr}
		\toprule
		          & Base & Temporal & TemporalR \\
		\midrule
		Base      & 1    & 0        & 0         \\
		Temporal  & 0    & 1        & 0         \\
		TemporalR & 0    & 0        & 1         \\
		\bottomrule
	\end{tabular}
	\caption{$p$ values of the post-hoc Conover test for the $M_{\ominus^5}$ scores on the TRG environment.}
		
\end{table}

\begin{table}[h]
	\centering
	\begin{tabular}{lrrrr}
		\toprule
		          & Base & Temporal & TemporalR \\
		\midrule
		Base      & 1    & 0        & 0         \\
		Temporal  & 0    & 1        & 0         \\
		TemporalR & 0    & 0        & 1         \\
		\bottomrule
	\end{tabular}
	\caption{$p$ values of the post-hoc Conover test for the $M_{\ominus^6}$ scores on the TRG environment.}
		
\end{table}

\begin{table}[h]
	\centering
	\begin{tabular}{lrrrr}
		\toprule
		          & Base & Temporal & TemporalR \\
		\midrule
		Base      & 1    & 0        & 0         \\
		Temporal  & 0    & 1        & 0.000006  \\
		TemporalR & 0    & 0.000006 & 1         \\
		\bottomrule
	\end{tabular}
	\caption{$p$ values of the post-hoc Conover test for the $M_{\ominus^7}$ scores on the TRG environment.}
		
\end{table}

\begin{table}[h]
	\centering
	\begin{tabular}{lrrrr}
		\toprule
		          & Base & Temporal & TemporalR \\
		\midrule
		Base      & 1    & 0        & 0         \\
		Temporal  & 0    & 1        & 0.000001  \\
		TemporalR & 0    & 0.000001 & 1         \\
		\bottomrule
	\end{tabular}
	\caption{$p$ values of the post-hoc Conover test for the $M_{\ominus^8}$ scores on the TRG environment.}
		
\end{table}

\begin{table}[h]
	\centering
	\begin{tabular}{lrrrr}
		\toprule
		          & Base & Temporal & TemporalR \\
		\midrule
		Base      & 1    & 0        & 0         \\
		Temporal  & 0    & 1        & 0.000077  \\
		TemporalR & 0    & 0.000077 & 1         \\
		\bottomrule
	\end{tabular}
	\caption{$p$ values of the post-hoc Conover test for the $M_{\ominus^1}$ scores on the TRG Hard environment.}
		
\end{table}

\begin{table}[h]
	\centering
	\begin{tabular}{lrrrr}
		\toprule
		          & Base & Temporal & TemporalR \\
		\midrule
		Base      & 1    & 0        & 0         \\
		Temporal  & 0    & 1        & 0         \\
		TemporalR & 0    & 0        & 1         \\
		\bottomrule
	\end{tabular}
	\caption{$p$ values of the post-hoc Conover test for the $M_{\ominus^2}$ scores on the TRG Hard environment.}
		
\end{table}

\begin{table}[h]
	\centering
	\begin{tabular}{lrrrr}
		\toprule
		          & Base & Temporal & TemporalR \\
		\midrule
		Base      & 1    & 0        & 0         \\
		Temporal  & 0    & 1        & 0         \\
		TemporalR & 0    & 0        & 1         \\
		\bottomrule
	\end{tabular}
	\caption{$p$ values of the post-hoc Conover test for the $M_{\ominus^3}$ scores on the TRG Hard environment.}
		
\end{table}

\begin{table}[h]
	\centering
	\begin{tabular}{lrrrr}
		\toprule
		          & Base & Temporal & TemporalR \\
		\midrule
		Base      & 1    & 0        & 0         \\
		Temporal  & 0    & 1        & 0         \\
		TemporalR & 0    & 0        & 1         \\
		\bottomrule
	\end{tabular}
	\caption{$p$ values of the post-hoc Conover test for the $M_{\ominus^4}$ scores on the TRG Hard environment.}
		
\end{table}

\begin{table}[h]
	\centering
	\begin{tabular}{lrrrr}
		\toprule
		          & Base & Temporal & TemporalR \\
		\midrule
		Base      & 1    & 0        & 0         \\
		Temporal  & 0    & 1        & 0         \\
		TemporalR & 0    & 0        & 1         \\
		\bottomrule
	\end{tabular}
	\caption{$p$ values of the post-hoc Conover test for the $M_{\ominus^5}$ scores on the TRG Hard environment.}
\end{table}

\begin{table}[h]
	\centering
	\begin{tabular}{lrrrr}
		\toprule
		          & Base & Temporal & TemporalR \\
		\midrule
		Base      & 1    & 0        & 0         \\
		Temporal  & 0    & 1        & 0         \\
		TemporalR & 0    & 0        & 1         \\
		\bottomrule
	\end{tabular}
	\caption{$p$ values of the post-hoc Conover test for the $M_{\ominus^6}$ scores on the TRG Hard environment.}
\end{table}

\begin{table}[h]
	\centering
	\begin{tabular}{lrrrr}
		\toprule
		          & Base & Temporal & TemporalR \\
		\midrule
		Base      & 1    & 0        & 0         \\
		Temporal  & 0    & 1        & 0.000041  \\
		TemporalR & 0    & 0.000041 & 1         \\
		\bottomrule
	\end{tabular}
	\caption{$p$ values of the post-hoc Conover test for the $M_{\ominus^7}$ scores on the TRG Hard environment.}
\end{table}

\begin{table}[h]
	\centering
	\begin{tabular}{lrrrr}
		\toprule
		          & Base & Temporal & TemporalR \\
		\midrule
		Base      & 1    & 0        & 0         \\
		Temporal  & 0    & 1        & 0.000119  \\
		TemporalR & 0    & 0.000119 & 1         \\
		\bottomrule
	\end{tabular}
	\caption{$p$ values of the post-hoc Conover test for the $M_{\ominus^8}$ scores on the TRG Hard environment.}
\end{table}

\end{document}